\documentclass[10pt,journal,compsoc]{IEEEtran}

\ifCLASSOPTIONcompsoc
  \usepackage[nocompress]{cite}
  \usepackage{url}
\else
  \usepackage{cite}
\fi

\ifCLASSINFOpdf
   \usepackage[pdftex]{graphicx}
\else
   \usepackage[dvips]{graphicx}
\fi

\usepackage{amsmath}
\usepackage{amsfonts}

\newcommand{\ours}{\textit{ISP}}
\newcommand{\chidb}{CHI3D}

\def\eg{\emph{e.g.}} 
\def\ie{\emph{i.e.}} 
\def\cf{\emph{c.f}.}

\newcommand{\bbeta}{\boldsymbol{\beta}}

\newcommand{\btheta}{\boldsymbol{\theta}}

\newcommand{\FF}{\mathbf{F}}
\newcommand{\FFH}{\mathbf{\hat{F}}}

\newcommand{\VV}{\mathbf{V}}
\newcommand{\VVH}{\mathbf{\hat{V}}}
\newcommand{\MM}{\mathbf{M}}
\newcommand{\MMH}{\mathbf{\hat{M}}}

\begin{document}
\title{Reconstructing Three-Dimensional Models of Interacting Humans}

\author{Mihai~Fieraru$^{1}$,
        Mihai~Zanfir$^{1}$,
        Elisabeta~Oneata$^{1}$,
        Alin-Ionut~Popa$^{1}$,
        Vlad~Olaru$^{1}$,
        and~Cristian~Sminchisescu$^{2,1}$%
\IEEEcompsocitemizethanks{\IEEEcompsocthanksitem $^{1}$Institute of Mathematics of the Romanian Academy.\protect\\
E-mail: \{firstname.lastname\}@imar.ro
\IEEEcompsocthanksitem $^{2}$Lund University.\protect\\
E-mail: cristian.sminchisescu@math.lth.se}%
}

\IEEEtitleabstractindextext{%
\begin{abstract}
Understanding 3d human interactions is fundamental for fine-grained scene analysis and behavioural modeling. However, most of the existing models predict incorrect, lifeless 3d estimates, that miss the subtle human contact aspects--the essence of the event--and are of little use for detailed behavioral understanding. This paper addresses such issues with several contributions: (1) we introduce models for interaction signature estimation (ISP) encompassing contact detection, segmentation, and 3d contact signature prediction; (2) we show how such components can be leveraged to ensure contact consistency during 3d reconstruction; (3) we construct several large datasets for learning and evaluating 3d contact prediction and reconstruction methods; specifically, we introduce CHI3D, a lab-based accurate 3d motion capture dataset with 631 sequences containing $2,525$ contact events, $728,664$ ground truth 3d poses, as well as FlickrCI3D, a dataset of $11,216$ images, with $14,081$ processed pairs of people, and $81,233$ facet-level surface correspondences. Finally, (4) we propose methodology for recovering the ground-truth pose and shape of interacting people in a controlled setup and (5) annotate all 3d interaction motions in CHI3D with textual descriptions. Motion data in multiple formats (GHUM and SMPLX parameters, Human3.6m 3d joints) is made available for research purposes at \url{https://ci3d.imar.ro}, together with an evaluation server and a public benchmark.
\end{abstract}

}

\maketitle

\IEEEdisplaynontitleabstractindextext

\IEEEpeerreviewmaketitle

\IEEEraisesectionheading{\section{Introduction}\label{sec:introduction}}

\begin{figure}[ht]
\begin{center}
        \includegraphics[width=.98\linewidth]{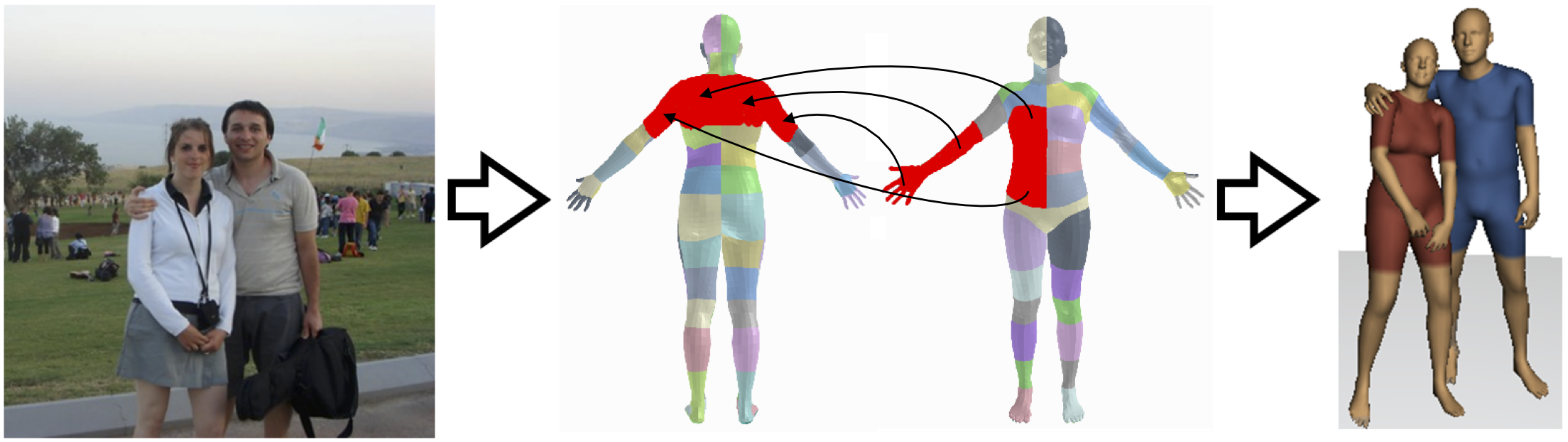}
\end{center}
\caption{Monocular 3d reconstruction, constrained by contact signatures, preserves the essence of the physical interaction between people and supports behavioral reasoning.}
\end{figure}

\IEEEPARstart{H}{uman}  sensing has recently seen a revival\cite{popa17,martinez17iccv,luvizon20182d,Yang_2018_CVPR} due to advances in large-scale deep learning architectures, powerful 3d kinematic and statistical shape models\cite{SMPL2015,Joo_2018_CVPR,Pavlakos2019}, as well as large scale 2d and 3d annotated datasets\cite{Lin14,CMUMotionCap2000,Ionescu14pami,Mahmood2019Amass}. While considerable progress has been achieved in localizing multiple humans in images, or reconstructing 3d humans in isolation, in a person-centred frame, little work has focused on inferring the pose and placement of multiple people in a three-dimensional, scene-centered coordinate system.

Moreover, the few approaches that have pursued such goals recently\cite{Mehta_3DV2018,Zanfir_2018_CVPR,zanfir_nips2018,li2019crowdpose,su2019multi,benzine2019deep,KolotourosIccv2019,VIBE:CVPR:2020}, concentrated mostly on the arguably difficult problem of multi-person data association, rather than the more subtle aspects such as close interactions during human contact. This leads to predictions that even when impressive in terms of plausible pose and shape from a distance, miss the essence of the event at close scrutiny, when \eg two reconstructions fail to capture the contact during a handshake, a tap on the shoulder, or a hug. Such interactions are particularly difficult to resolve as effects compound: on one hand depth and body shape uncertainty could result in compensation by pushing limbs in front or further away from their ground truth position, when inferring 3d from monocular images\cite{sminchisescu_cvpr03}; on the other hand, partial occlusion and the relatively scarce detail (resolution) for contact areas in images, typical of many human interactions, can make visual evidence inconclusive. %

In this paper, we %
propose a first set of methodological elements to address the reconstruction of interacting humans, in a more principled manner, %
by relying on recognition, segmentation, mapping, and 3d reconstruction. More precisely, 
we break down the problem of producing veridical 3d reconstructions of interacting humans into (a) contact detection, (b) binary segmentation of contact regions on the corresponding surfaces associated to the interacting people; (c) contact signature prediction to produce estimates of the potential many-to-many correspondence map between regions in contact; and (d) 3d reconstruction under augmented losses built using additional surface contact constraints given a contact signature. To train models and evaluate the techniques we introduce two large datasets: \chidb{}, a lab-based 3d motion capture repository containing $631$ sequences containing $2,525$ contact events, 
$728,664$ ground truth skeletons, as well as FlickrCI3D, a dataset of $11,216$ images, with $14,081$ processed pairs of people, and $81,233$ facet-level surface contact correspondences. 
In extensive experiments, we evaluate all system components and provide quantitative and qualitative comparisons showing how the proposed approach can %
capture 3d human interactions realistically. 

An earlier version of this manuscript appeared in \cite{Fieraru_2020_CVPR}. This version adds several new contributions. First, we annotate the whole temporal extent of the physical contact in each video sequence. Second, in sec.~\ref{sec:chi3d_gt}, we propose methodology to obtain the ground truth pose and shape of the interacting people in the CHI3D dataset. This goes beyond lifting marker 3d positions to the parameters of a body model as: (a) only one subject in each video is motion tracked (a limitation we inherit from the Vicon motion capture system we use) and 
(b) our subjects are not only in close proximity, but they are also often in physical contact. By leveraging information from motion sensors, multi-view RGB cameras and a 3d scanner, but also from contact annotations, pose priors and physical constraints, we achieve ground-truth level 3d reconstructions (see fig.~\ref{fig:chi3d_fitting}). 
Third, we annotate each CHI3D interaction motion with a textual description, which opens the door for training and evaluating models that generate 3d interaction motions from text.
Fourth, in addition to the 3d joint sequences, we publicly release the ground-truth motion sequences in both GHUM \cite{xu2020ghum} and SMPLX \cite{Pavlakos2019} format (see \url{ci3d.imar.ro}) and implement an evaluation server on a hidden test set, together with a public benchmark, with the purpose of advancing the state of the art in 3D human reconstruction of close contact interactions. The evaluation protocol and its motivation are presented in sec.~\ref{sec:benchmark}, together with initial benchmark results of popular monocular 3d reconstruction methods.

\section{Related Work}
\noindent{\bf Human and Object Interactions.} 
Most previous work addressed the topic of interactions without explicit modeling of the person-to-person or person-to-object contact. For instance, the 3d reconstruction of multiple people in the context of close interactions was partially addressed in \cite{liu2013markerless,liu2011markerless}, where 3d human skeletons were rigged to mesh surfaces of participants. 
Scenes were captured using a multi camera setup on a green background. 3d pose estimation and shape modelling were performed using energy-based optimization, taking into account the multi-camera setup, green background separation and temporal consistency. Human interactions or contact were not modeled explicitly beyond non-penetration of mesh surfaces.
Yun et al.\cite{yun2012two} proposed methods for action classification in scenes with two interacting people using RGB-D data and multiple instance learning. However, their data does not imply physical interaction between subjects and no form of contact is labeled. Hand to hand interaction is studied in \cite{Tzionas2016,GCPR_2013_Tzionas_Gall,Taylor2017,mueller_siggraph2019}, where models are optimized using energy minimization with non-penetration constraints but without a contact model. Other methods focus on the interaction between the 3d human shape and the surrounding environment\cite{Hamer2009,Oikonomidis2011,ballan2012,Tzionas2016,Pham2018,Hasssan2019Prox,brahmbhatt2019contactdb}, in most cases without a detailed object contact model.
Recently, REMIPS \cite{NEURIPS2021_a1a2c3fe} diverged from the optimization based reconstruction proposed in this work and proposed using the contact signatures (both interaction and self-contact) as weak supervision of an end-to-end trained network.  PI-Net \cite{guo2021pi} also estimates the 3d poses of interacting people, but predicts only their 3d keypoints and does not show results on the challenging cases of interaction contact.\\

\noindent{\bf Self-contact} The previous version of this manuscript \cite{Fieraru_2020_CVPR} also inspired the study of human self-contact and its use in 3d human pose and shape estimation. \cite{fieraru2021learning} proposed predicting self-contact signatures together with self-contact positioning in image space and showed the benefits of constraining 3d human pose and shape optimization with self-contact signatures. Following the same contact signature definition, \cite{Muller_2021_CVPR} also leverages self-contact annotations in combination with several types of data (3d scans, mimic-the-pose images, in-the-wild images) to improve 3d human pose and shape estimation. \\

\noindent{\bf Psycho-social Studies.} \cite{Suvilehto13811} construct a body region map to describe the most likely contact areas for different types of social bonds (\eg child and parent, siblings, life partners, casual friends, strangers) and conclude that social contact between two individuals varies with emotional bondage. Human close interaction analysis could be important in social studies involving robot assisted therapy for autistic children. \cite{marinoiu18deenigma,rudovic2017measuring} record robot assisted therapy sessions and perform extensive analysis over the interactions between the therapist, the children and the robot. 
Research in this area can impact social group interaction analysis \cf \cite{salam2016fully,celiktutan2017multimodal}. A similar study \cite{leclere2016} was performed over the dyadic relationship between mother and children.\\

\noindent{\bf Datasets.}  Most datasets dedicated to human understanding are centered around single person scenarios\cite{CMUMotionCap2000,Ionescu14pami,sigal2010,Mahmood2019Amass} and even those that include multiple people \cite{Lin14,vonMarcard2018,Everingham10} do not explicitly model the close interaction between different people. 
Human interactions have been captured before in classification/recognition datasets, one such example being \cite{gu2018ava}. The dataset provides action labels for short video sequences (\ie under $1$-$2$ seconds) collected from  YouTube. However, the dataset does not provide detailed contact annotations either in the image or at the level of 3d surfaces, as pursued here. Following our initial work \cite{Fieraru_2020_CVPR}, two more datasets focused on close human interactions were introduced. The ExPI dataset \cite{guo2022multi} contains the motion and shape of 2 interacting people in a controlled setup. The type of actions are, however, limited to Lindy-hop dancing actions and the motion is provided only in the 3d joints format, not also in the format of a parametric body model (e.g. GHUM \cite{xu2020ghum} or SMPLX \cite{Pavlakos2019}). A very recent work \cite{yin2023hi4d} is also introducing a dataset focused on close human interactions called Hi4D. Similar to us, they reconstruct the interactions in a controlled setup and extract vertex-level contact signatures. They, however, do not model hands, which are crucial in most of the interactions between humans. CHI3D is modelling and fitting hands by using expressive full body models \cite{xu2020ghum, Pavlakos2019}, hand priors \cite{zanfir2020weakly} and human annotated contact correspondences.

\section{Datasets and Annotation Protocols}

\noindent{\bf FlickrCI3D.} We collect images from the YFCC100M dataset\cite{thomee2016yfcc100m}, a database containing photos uploaded to Flickr by amateur photographers who share their work under a Creative Commons license. Using \cite{kalkowski2015real} to search the dataset, we download images expected to contain scenes with close interactions between people. We either query the dataset using tags generic to the human category, such as ``persons``, ``friends``, ``men``, ``women``, or using tags related to actions performed by humans in physical contact, such as ``dance``, ``hug``, ``arrest``, ``handshake``.  We run a common multi-person 2d keypoint estimator\cite{cao2018openpose} - to detect the humans in each picture and select all pairs of people whose bounding boxes overlap. We automatically filter out the images with small resolution, where pairs of people are severely occluded or have large scale differences. We refer to this data collection together with the underlying 3d surface contact annotations (described in the sequel) as FlickrCI3D.\\

\noindent{\bf \chidb{}.} We also collect a lab-based 3d motion capture dataset, \chidb{} (\underline{C}lose \underline{H}uman \underline{I}nteractions 3D), for quantitative evaluation and possibly training %
of 3d pose and shape reconstruction. 
We employ a Vicon MoCap system of $10$ motion cameras synchronized with $4$ additional RGB cameras to capture short video sequences of $6$ human subjects, grouped in $3$ pairs, performing close interaction scenarios: grab, handshake, hit, holding hands, hug, kick, posing (for picture) and push. 
Since the MoCap system setup prohibits %
the use of two sets of markers on both people at the same time during a recording, 
each of the human subjects takes turns on wearing the body markers. 
In total, we collect $631$ sequences consisting of %
$485,776$ pairs of RGB frames and MoCap skeleton configurations.

\subsection{Annotation Protocol}
We next describe the manual annotation pipeline, which consists of two stages. First, %
the annotators label the existence of a contact between two people.
Second, they localize the physical contact for pairs of people who were previously annotated to be in contact, by establishing correspondences between two 3d human body meshes. In both steps, the annotators are presented with a picture and two superimposed 2d skeletons identifying the people of interest. This approach helps clear the identity confusion in crowded scenes or interactions with high people overlap.\\

\begin{figure}[!htbp]
\begin{center}
         \includegraphics[height=63pt]{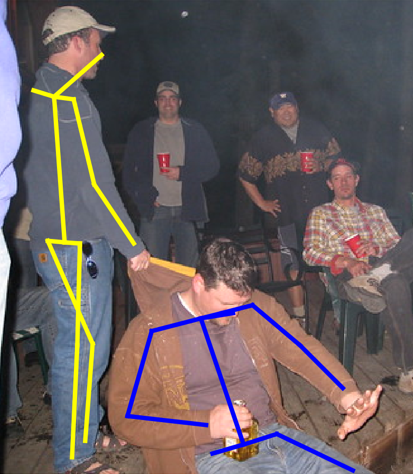}
         \includegraphics[height=63pt]{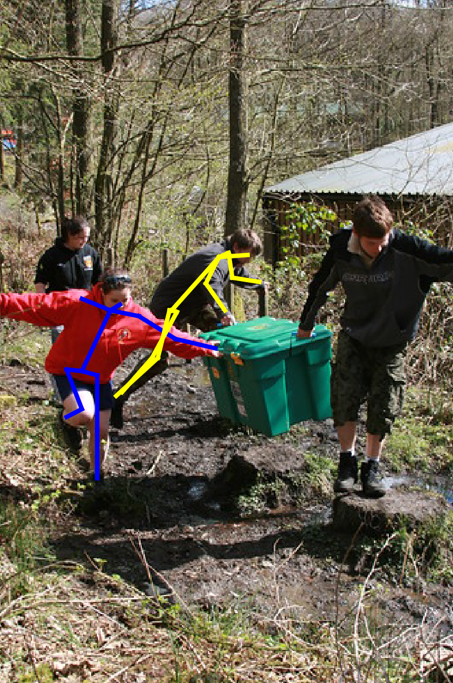}
         \includegraphics[height=63pt]{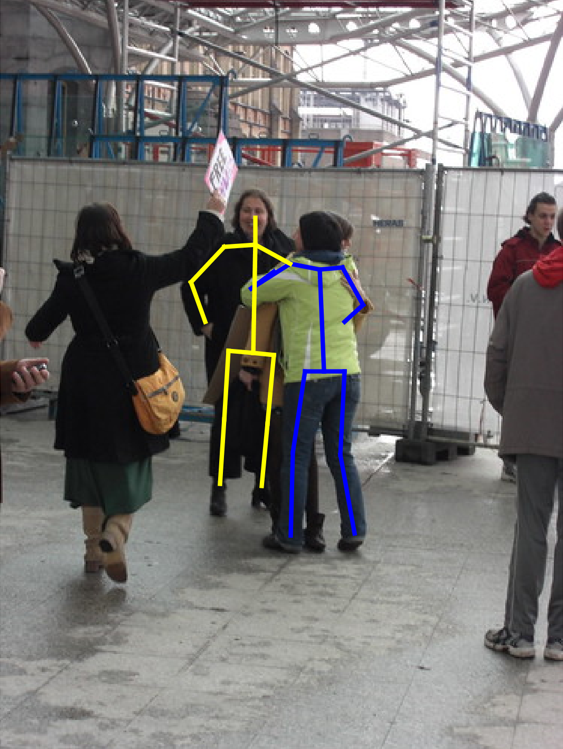}
         \includegraphics[height=63pt]{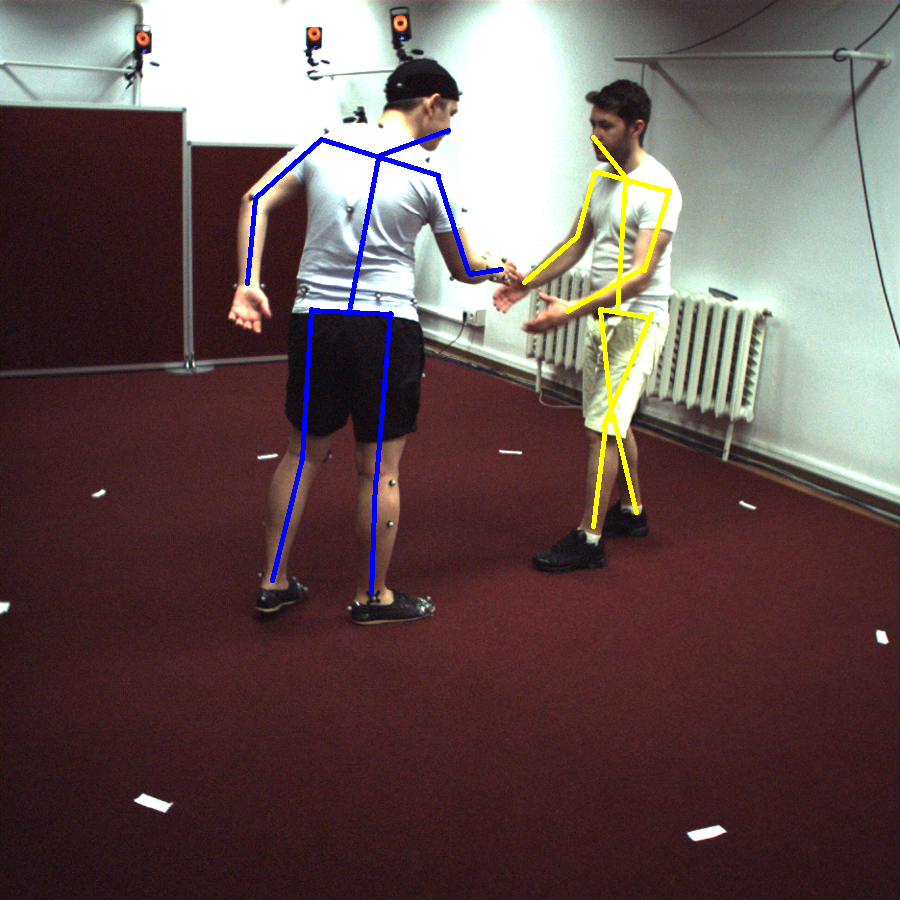}
         \\
         \includegraphics[height=67pt]{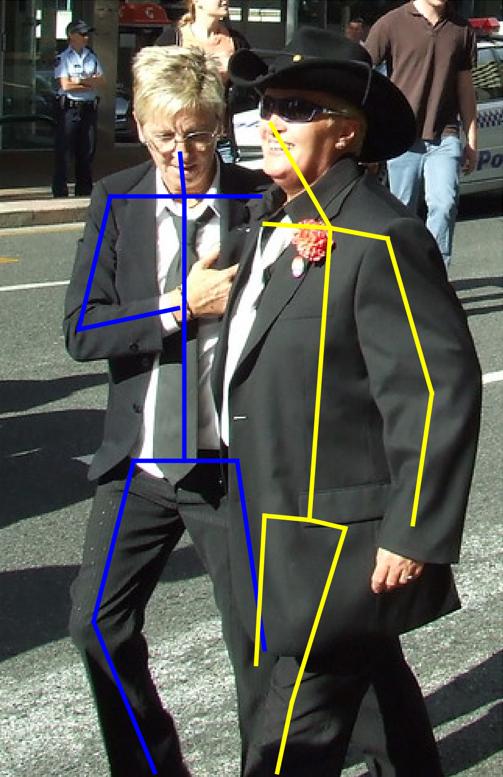}
         \includegraphics[height=67pt]{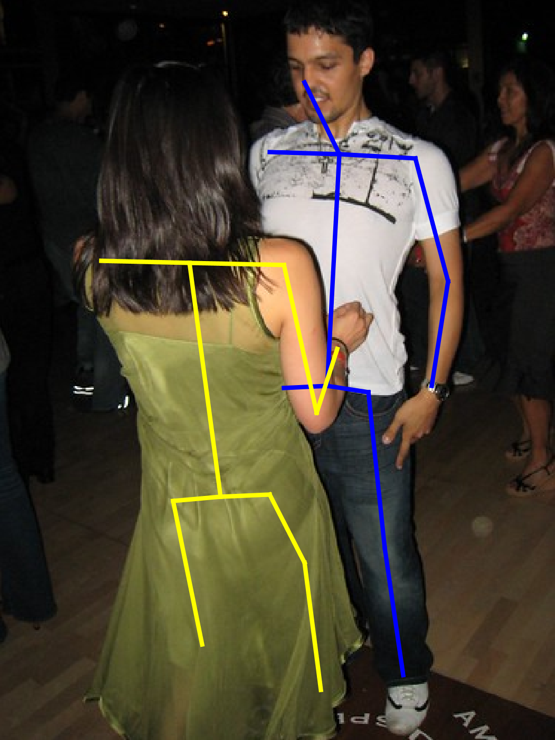}
         \includegraphics[height=67pt]{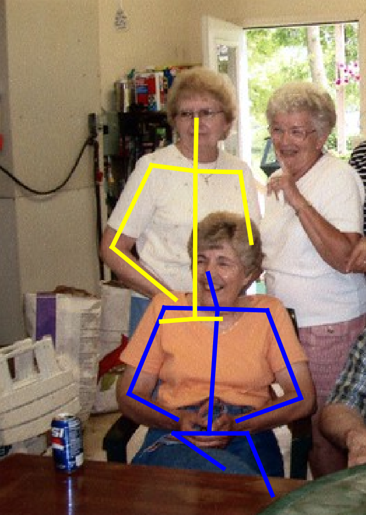}
        \includegraphics[height=67pt]{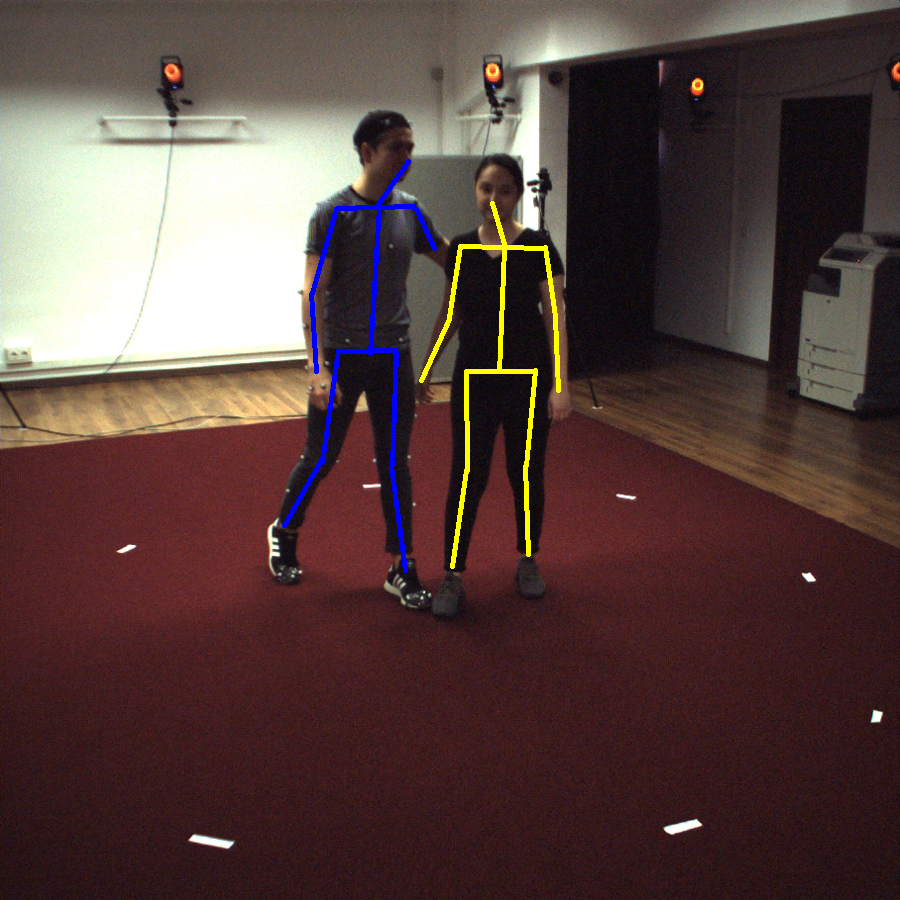}
         \\
         \includegraphics[height=61pt]{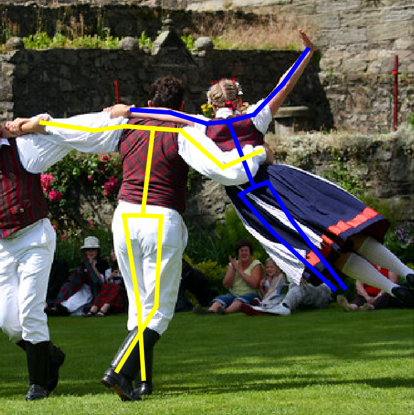}
         \includegraphics[height=61pt]{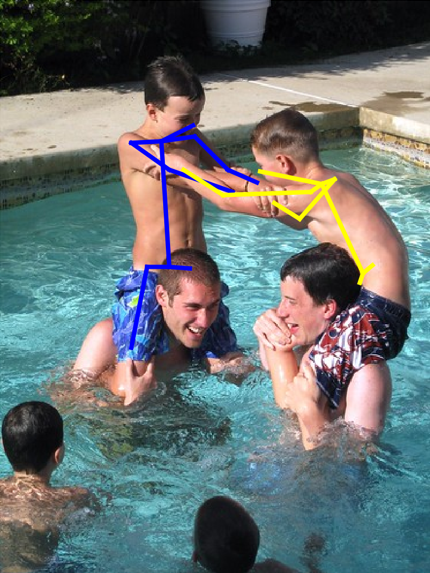}
         \includegraphics[height=61pt]{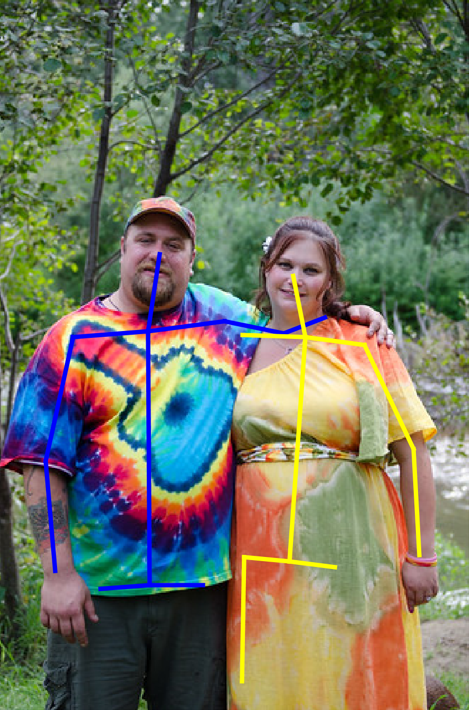}
         \includegraphics[height=61pt]{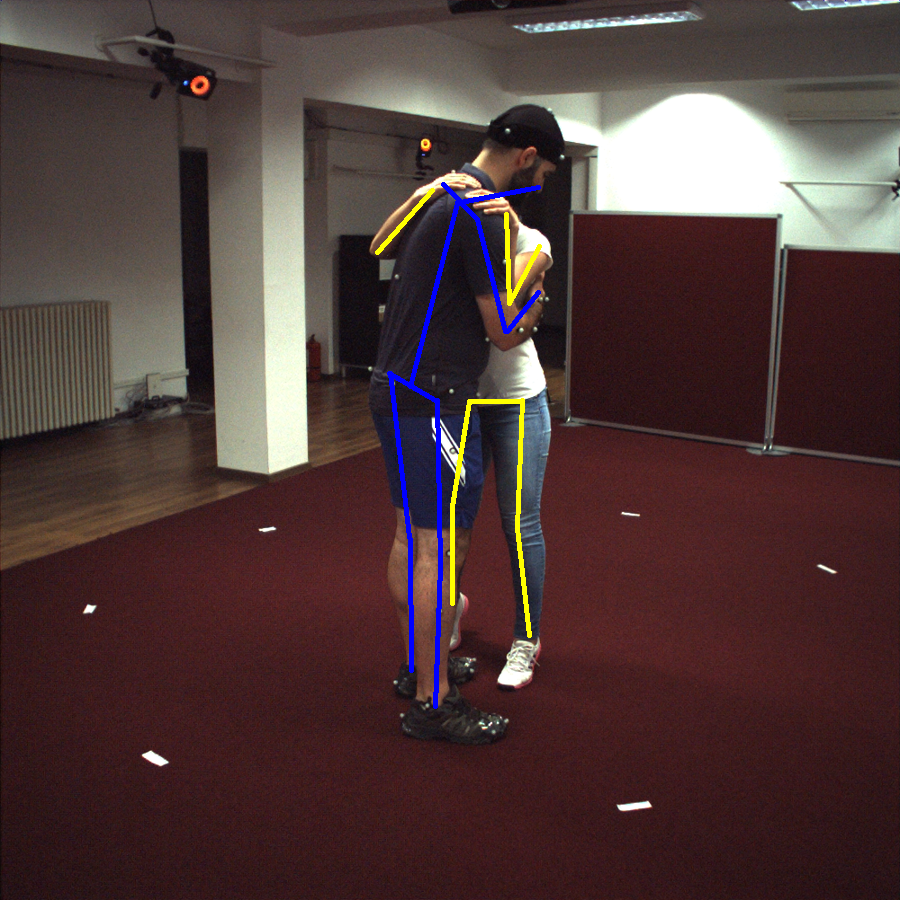}
\end{center}
\caption{Contact classes examples in FlickrCI3D and \chidb{} (last column). \textbf{First Row}: "no contact", %
clearly visible that the two people are not touching at all. \textbf{Second Row}: "uncertain contact", %
there is ambiguity if there is contact or not. \textbf{Third Row}: "contact", %
the contact between the two persons is clearly visible.}
\label{fig:flickr_examples}
\end{figure}

\noindent{\bf Contact Classification.} Given a scene where the detector identified 2d body poses in close proximity, we identify four scenarios that have to be manually classified: (1) \textbf{erroneous 2d pose estimations}, \ie the assignment between the estimated skeletons and the people in the image cannot be determined, or at least two estimated limbs have no overlap with the real limbs in the image, (2) certainly \textbf {no contact} between the two people, (3) \textbf{contact} between the two people and (4) \textbf{uncertain contact} between the two people, \ie both "contact" and "no contact" cases may be possible, but it is ambiguous in the given image.
On FlickrCI3D, annotators are instructed to label each pair of people with one of these four classes, which is achieved at an annotation rate of around $500$ pairs / hour. By discarding the few pairs of 2d skeletons that are erroneous ($8\%$), the result is a database of $65,457$ images, containing $90,167$ pairs of people in close proximity, in the following proportions: $18\%$ "contact", $21\%$ "uncertain contact", $61\%$ "no contact". Example images from each of these classes are shown in fig.~\ref{fig:flickr_examples}. 
On \chidb{} video sequences, annotators are instructed to select the entire temporal extent of the physical contact. Since more information is available, we show annotators all $4$ video views.  \\

\noindent{\bf 3D Contact Signature Annotation.}
When two people are in physical contact, we want to understand \textit{where} and \textit{how} they interact by encoding the information on the surfaces of two 3d human meshes. 

To this end, we define the \textbf{facet-level contact signature} $C^{facet}(I, P_1, P_2) \in \{0, 1\}^{N_{facets} \times N_{facets}}$ between two people $P_1, P_2$ in image $I$ as $C^{facet}_{f_1, f_2}(I, P_1, P_2)=1$ if facet $f_1$ of the mesh of person $P_1$ is in contact with facet $f_2$ of the mesh of person $P_2$ and $C^{facet}_{f_1, f_2}(I, P_1, P_2)=0$ if they are not in contact. 

We also define the \textbf{facet-level contact segmentation} $S^{facet}(I, P_1, P_2) \in \{0, 1\}^{N_{facets} \times 2}$ of the contact of two people $P_1, P_2$ in image $I$ as $S^{facet}_{f, i}(I, P_1, P_2)=1$ if facet $f$ of the mesh of person $P_i$ is in contact with any other facet of the mesh of the other person, and $S^{facet}_{f, i}(I, P_1, P_2)=0$ otherwise. Note that the contact segmentation $S$ can be recovered from the contact signature $C$.

State of the art body meshes \cite{xu2020ghum,Pavlakos2019} have a large number of surface facets, $N_{facets}\approx20,000$. Annotating a contact signature with high fidelity in such a huge dimensional space, \ie $N_{facets} \times N_{facets}$, is both tedious and time-consuming.

\begin{figure}[!htbp]
\centering         
         \includegraphics[width=0.8\linewidth]{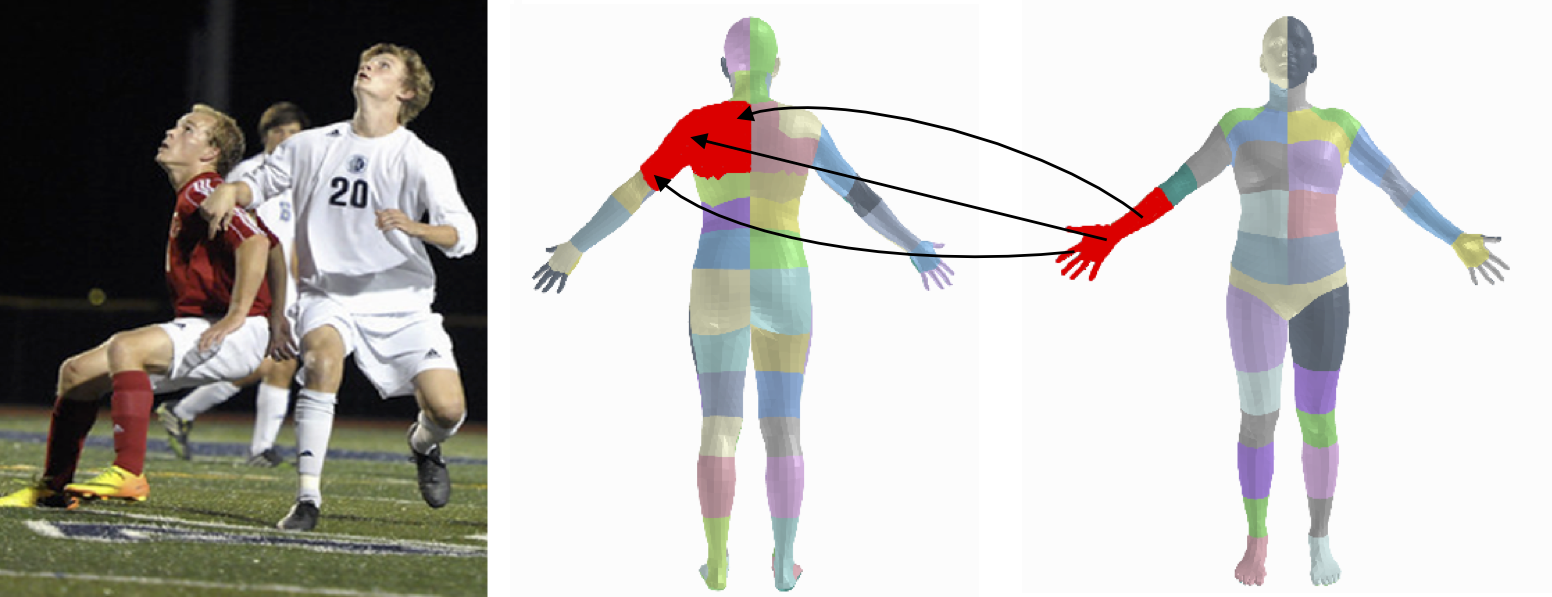}
         \\
         \includegraphics[width=0.8\linewidth]{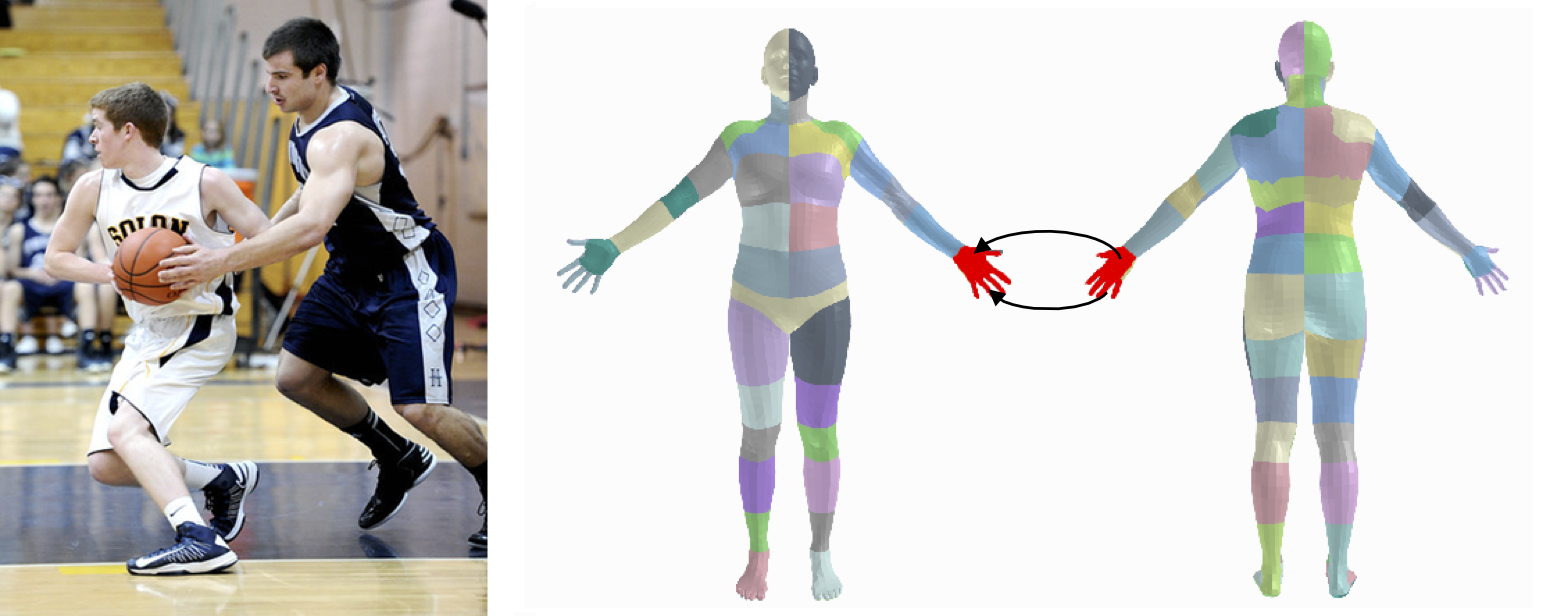}
\caption{3d contact segmentations and signatures from FlickrCI3D. For an RGB image, the annotators map facets from one mesh to facets on the other mesh if they are in direct contact. By doing so, they automatically segment the regions in contact (marked in red) and facet-level correspondences (marked by arrows).}
\label{fig:flickr_correspondences}
\end{figure}

An alternative to simplify the annotation burden is to first perform %
contact segmentation and then establish correspondences only between the contact segments on both surfaces. %
However, even fine segmentation annotation of the contact on 3d surfaces is cumbersome and requires a high degree of precision and creativity. 
Instead, we relax the annotation granularity and group the $N_{facets}$ facets into a number of $N_{reg}=75$ predefined regions. We guide our grouping strategy by following the anatomical parts of the human body and their symmetries, as seen in fig.~\ref{fig:flickr_correspondences}.
\begin{table}
\setlength{\tabcolsep}{0.8em} 
\begin{center}
\begin{tabular}{|c|cc|cc|}
\hline
 & \multicolumn{2}{c|}{Segmentation IoU}  & \multicolumn{2}{c|}{Signature IoU}  \\
\# Reg. & \chidb{} &  FlickrCI3D &  \chidb{} &  FlickrCI3D \\
\hline\hline
$75$ & $0.692$ & $0.456$ & $0.472$ & $0.226$ \\
$37$ & $0.790$ & $0.542$ & $0.682$ & $0.370$ \\
$17$ & $0.815$ & $0.638$ & $0.721$ & $0.499$ \\
$9$ & $0.878$ & $0.745$ & $0.799$ & $0.635$ \\
\hline
\end{tabular}
\end{center}
\caption{Annotator consistency as a function of the granularity of surface regions. The task has an underlying ground truth, but it is sometimes hard for annotators to identify it. At 17 and 9 regions partitioning, respectively, there is reasonable consistency. Notice that for \chidb{} the consistency is higher as the annotators rely on 4 views of the contact. }
\label{table:consistency}
\end{table}

Now, the definitions of \textbf{region-level contact signature} $C^{reg}(I, P_1, P_2) \in \{0, 1\}^{N_{reg} \times N_{reg}}$ and \textbf{region-level contact segmentation} $S^{reg}(I, P_1, P_2) \in \{0, 1\}^{N_{reg} \times 2}$ follow naturally by considering two regions $r_1$ and $r_2$ to be in contact if at least one facet from region $r_1$ is in contact with at least one facet from region $r_2$. With such a setup, segmentation can be performed quickly with a few clicks on the regions in contact. 

To support the annotation effort, we implemented a custom 3d annotation interface which displays an image, the superimposed 2d poses of the people of interest, alongside two rendered 3d body meshes that can be manipulated via rotations, translations or zoom. %
Each facet-level correspondence is annotated one at a time, by
choosing one facet on each surface. The regions containing the two facets are automatically colored in red to illustrate that they are now segmented as contact regions. The annotators proceed labeling other correspondences, until the region-level contact segmentation is complete. The choice of which facet-level correspondences to label is up to the annotators. Using this simplified annotation process does not guarantee a complete set of correspondences. %
 The annotators accomplish a rate of around 25 pairs of people / hour. Some examples of annotations are shown in fig.~\ref{fig:flickr_correspondences}.

Note that while in \chidb{} the annotators are shown all 4 views of the contact scene, in FlickrCI3D they have access to only one view of the interaction. Although the pairs of people are certainly in contact (as annotated in the %
first stage), there can still be ambiguity on the precise configuration of the contact, mostly caused by occlusions. In such cases, we instruct annotators to imagine one possible configuration of the contact signature and annotate it.\\

\noindent{\bf Textual Annotations.} After 3d reconstructing the interactions in CHI3D (see sec.~\ref{sec:chi3d_gt}), we annotate each pair of mesh sequences with a text describing the interaction motion. The $631$ text annotations (one per interaction) average $12$ English words and support further augmentations (e.g. changing the order of the motions of the two people without changing the textual annotation). This is a first step towards learning models that can generate 3d human motion interactions from text. 
 
\noindent{\bf Datasets Statistics.}
As the signature annotation task is expensive and time-consuming, each interaction in the datasets is labeled by only one annotator. Following the annotation process on FlickrCI3D, we gather a number of $11,216$ images and $14,081$ valid pairs of people in contact, with
$81,233$ facet-level correspondences within $138,213$ selected regions. This results in an average of $5.77$ correspondences per pair of people. For \chidb{}, annotators select one contact frame per sequence for contact signature labeling, which results in a total of $2,524$ images and 
pairs of people, with $10,168$ facet-level correspondences within $15,168$ selected regions. This results in an average of $4.03$ correspondences per pair of people. 

\begin{figure}[!htbp]
\begin{center}
         \includegraphics[width=\linewidth]{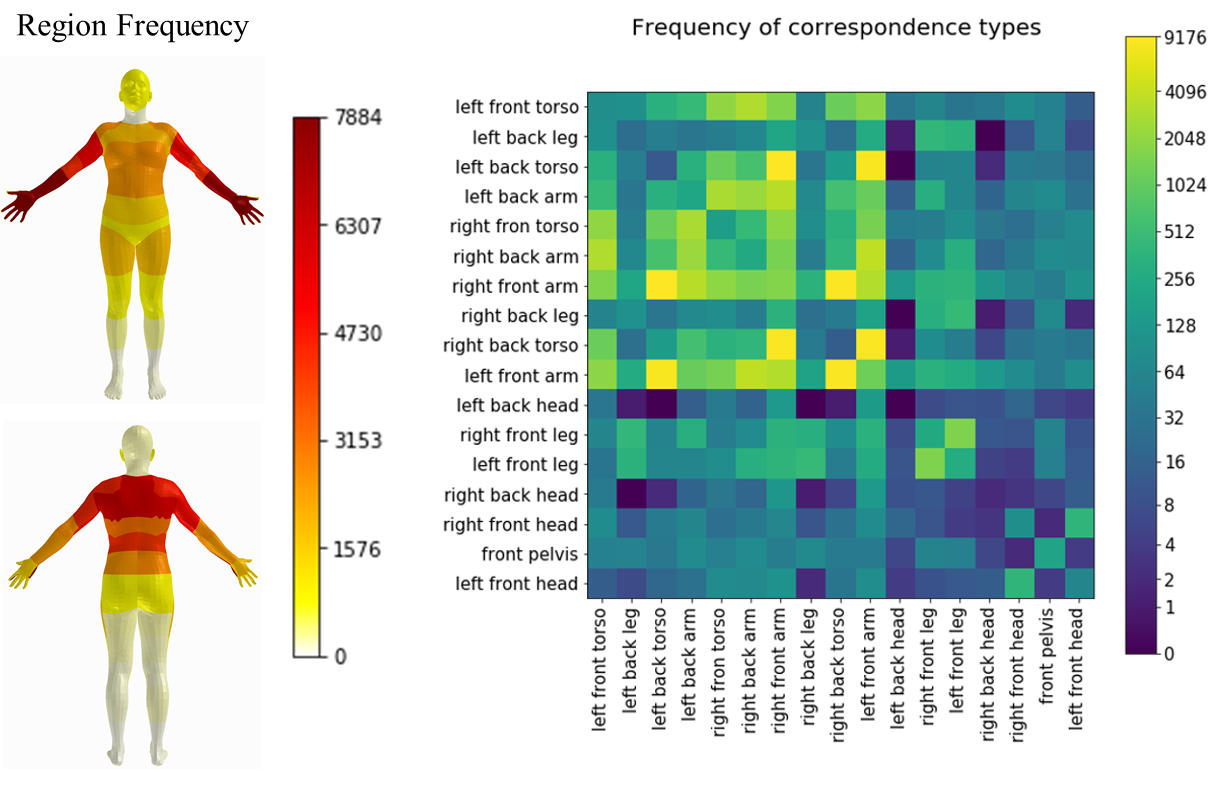}
\end{center}
\caption{(\textit{\textbf{Left}}) Frequency of body regions involved in a contact (75 regions). Note the left-right symmetry and the high frequency for the arms, shoulders and back regions. (\textit{\textbf{Right}}) Correspondence frequency counts (17 regions).}
\label{fig:clicks_frequencies}
\end{figure}

\begin{figure*}[!htbp]

\begin{center}
         \includegraphics[width=0.99\linewidth]{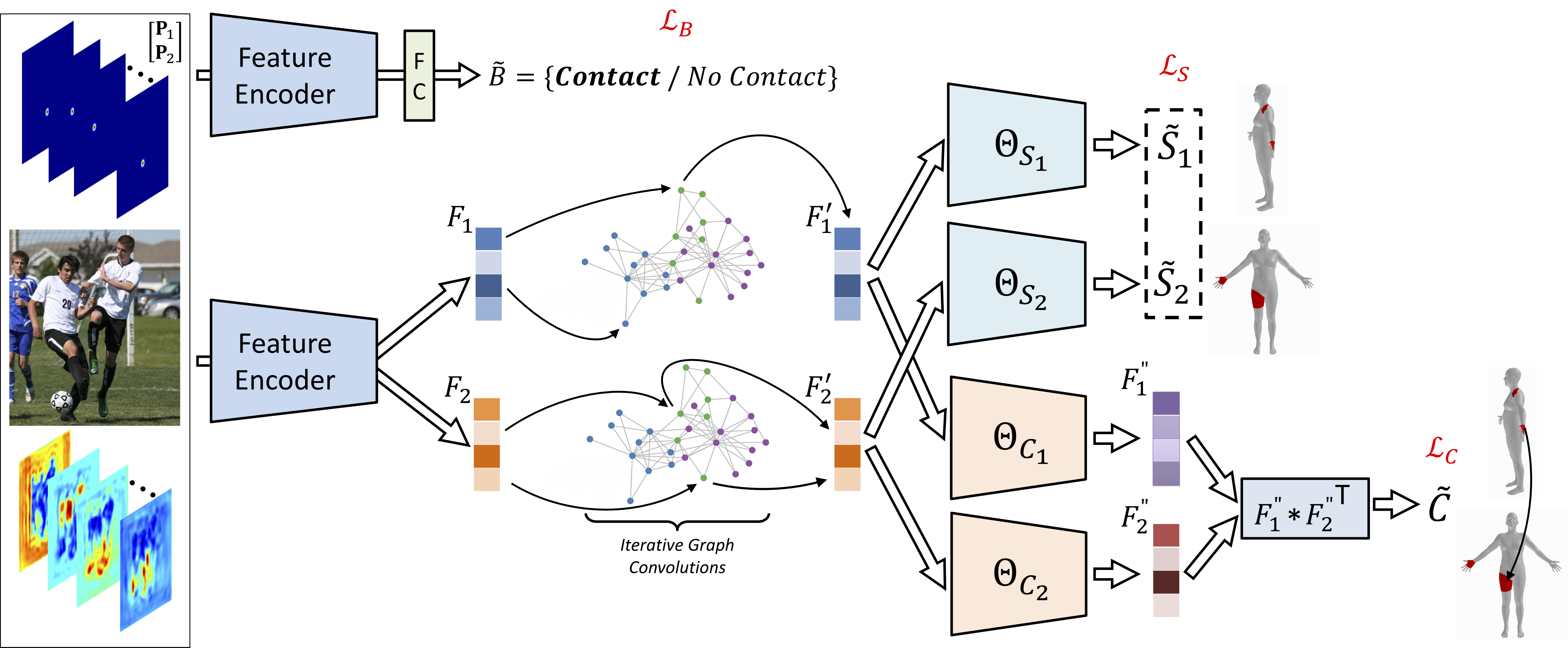}
\end{center}
\caption{Multi-task architecture \ours{} for detection and prediction of interaction signatures, %
that (1) classifies whether people are in contact, (2) holistically segments the corresponding 3d body surface contact regions for each person, and (3) determines their specific 3d body contact signature. %
Each task has a specific loss, ${\mathcal{L}}_B$, ${\mathcal{L}}_S$ and ${\mathcal{L}}_C$ respectively. As input, we feed an RGB image of people in close proximity, two 2d skeleton predictions and semantic human features computed on the image. The binary contact estimation task uses a single fully connected layer. The 3d contact segmentation and signature prediction tasks use a sequence of fully connected layers and graph convolutions (shared by both tasks), followed by fully connected layers specialized for each task.
}
\label{fig:pipeline_overview}
\end{figure*}
In fig.~\ref{fig:clicks_frequencies} (left) we show a 3d human region heat map based on the frequencies of the regions involved in a contact. 
Notice that the front side of the hands/arms as well as the back side of a person are the most common body parts involved in contacts. This observation is also %
confirmed by the work of \cite{Suvilehto13811} who give contact region maps for various types of human relationships. In fig.~\ref{fig:clicks_frequencies} (right) we show a frequency map of the correspondences at region level. Notice the large coverage of annotated contact correspondences.

Given the ambiguities of determining contact correspondences from a single view, we check the consistency between annotators on a small common set of images. Results can be seen in Table ~\ref{table:consistency}. We evaluate the consistency considering different levels of granularity when grouping facets into regions. It can be noticed that at the highest level of detail they have lower consistency, but as coarser regions of the body are aggregated, consistency increases. This observation is partly congruent to the computational perception study of \cite{marinoiuijcv16} who argue that humans are not very precise in re-enacting 3d body poses viewed in monocular images. Note that, for \chidb{}, consistency is higher as multiple views are available in the annotation process.

\section{Methodology}
In this section, we describe the models we introduce for the following tasks: (1) contact detection (classification), (2) 3d contact segmentation, (3) 3d contact signature prediction and (4) 3d pose and shape reconstruction using contact signatures.

For tasks (1)-(3), we propose learning methods (collectively referred as \emph{Interaction Signature Prediction - ISP}) based on deep neural networks that take as input an image $I$ cropped around the bounding box of the two interacting people $P_1$, $P_2$, together with the associated 2d human body poses detected\cite{cao2018openpose}. We encode each 2d body pose as $n_{kp}$ channels, one for each keypoint type, by considering a 2d Gaussian around the coordinate of each keypoint. Following \cite{fieraru2018learning}, we stack the two pose encodings with the RGB image $I$. In addition, we also stack semantic human features to the input, \ie 2d body part labeling\cite{popa17} and 2d part affinity fields, both computed on $I$. 

We use the ResNet50\cite{he2016deep} backbone architecture (up to the last average pooling layer) as a trainable feature encoder, which we modify to accommodate the larger number of input channels by increasing the size of the first convolutional filters. An overview of the pipeline is given in fig.~\ref{fig:pipeline_overview}.
\subsection{Contact Classification \label{sec:met_classif}}
Given an image $I$ with two people in close proximity we want to estimate if there is \textit{any} physical contact between the two. 
We train a deep binary classification network composed of the feature encoder network to which we add a fully connected layer which outputs the probability of the two
label classes, $B = \{0, 1\}$, $1$ -- "contact" and  $0$ -- "no contact". We train using the weighted binary cross entropy loss function
with $w_0 < w_1$ as the weights for classes $0$ and $1$, respectively, to account for the more frequent "no-contact" class. 
\begin{table*}[!htbp]
\begin{center}
\scalebox{0.88}{
\begin{tabular}{|c|cc|cc|cc|cc|}\hline
& \multicolumn{2}{c}{\boldmath{$\text{IoU}_{75}$}} \vline &
\multicolumn{2}{c}{\boldmath{$\text{IoU}_{37}$}} \vline &
\multicolumn{2}{c}{\boldmath{$\text{IoU}_{17}$}} \vline &
\multicolumn{2}{c}{\boldmath{$\text{IoU}_{9}$}} \vline \\

\textbf{Method} 
& \textbf{Segm.} & \textbf{Signature} & \textbf{Segm.} & \textbf{Signature}  & \textbf{Segm.} & \textbf{Signature} & \textbf{Segm.} & \textbf{Signature}\\
\hline
\hline
\ours{} full  & $\mathbf{0.318}$ & $\mathbf{0.082}$ & $\mathbf{0.365}$ & $\mathbf{0.129}$ &  $\mathbf{0.475}$ & $\mathbf{0.248}$  & $\mathbf{0.618}$ & $0.408$\\
\ours{} w/o semantic 2d features as input & $0.300$ & $0.073$ & $0.350$ & $0.116$ & $0.465$ & $0.240$ & $\mathbf{0.618}$ & $\mathbf{0.410}$\\
\ours{} w/o jointly learning contact segm. & - & $0.072$ & - & $0.124$ & - & $0.218$ & - & $0.383$ \\
\ours{} w/o masking out corresp. outside the estimated segm. mask & - & $0.075$ & - & $0.124$ & - & $0.230$ & - &$0.385$ \\
\hline
Human performance  & $0.456$ & $0.226$ & $0.542$ & $0.370$ & $0.638$ & $0.499$ & $0.745$ & $0.635$ \\
\hline
\end{tabular}
}
\end{center}
\caption{Results of our contact segmentation and signature estimation on FlickrCI3D, evaluated for different region granularities on the human 3d surface (from 75, down to 9 regions). We ablate different components of our full method to illustrate their contribution. Human performance represents the consistency values between annotators from table~\ref{table:consistency}.}
\label{table:contact3d_quant}
\end{table*}

\subsection{Contact Segmentation and Signature\label{sec:met_segm_corresp}}

In order to operate within a manageable output space, we consider contact segmentation and signature prediction at region-level. In the following, let $N = N_{reg}$, $S = S^{reg} \in \mathbb{R}^{N\times 2}$, the ground-truth contact segmentation, and $C =  C^{reg}  \in \mathbb{R}^{N\times N}$, the ground-truth contact signature.
We leverage the synergy between the segmentation and signature tasks and 
train them together in a multi-task setting.

 Following the feature encoder backbone, we split the network into separate computational pathways for each person, in order to better disentangle their feature representations. As a first step, we extract two sets of features $F_p \in \mathbb{R}^{N\times D_{0}}$, 
 $p=1,2$, using fully connected layers, where $D_{0}$ is the size of the features. To integrate the topology of the regions on the mesh, we next apply a fixed number of graph convolution iterations, following the architecture proposed in \cite{kolotouros2019cmr}. The adjacency matrix we use corresponds to the $N$ regions on the 3d template body mesh, where we set an edge between two regions if they share a boundary. We denote by $F_p^{'}$ the output of the graph convolutions. We pass these features to segmentation ($\Theta_{S_p}$) and signature ($\Theta_{C_p}$) specialization layers, each implemented as a fully connected layer. 

The output of the $\Theta_{S_p}$ layers, $\widetilde{S} = [\widetilde{S}_1  \widetilde{S}_2]$, represents the final segmentation prediction for the two persons. We use the sigmoid cross-entropy loss 
\begin{equation}
    \mathcal{L}_{S}(I) =  -\sum_{i=1}^{2 \times N}{(p_{S}S_{i}  \log(\widetilde{S}_{i}) + (1 - S_{i}) \log(1 -\widetilde{S}_{i}))}
\label{egn:crossentropy}
\end{equation}
with a balancing term $p_{S} \in \mathbb{R}$ between the positive and negative classes.
For the contact signature prediction task, we use the output of the $\Theta_{C_p}$ layers, $F_p^{''}$, and compute our estimate as $\widetilde{C} = F_1^{''} * F_2^{''T}$. %
We again use the cross entropy loss, $\mathcal{L}_C$, with the difference that we iterate to $N \times N$ and use another balancing term, $p_C \in \mathbb{R}$. 

\subsection{Monocular 3D Reconstruction\label{sec:monoc_rec}}
Given an image $I$ and its contact signature $C(P_1, P_2)$, we want to recover the 3d pose and shape parameters of the two people in contact, $P_1$ and $P_2$. We start from the optimization framework of \cite{Zanfir_2018_CVPR} and augment it with new loss terms that explicitly use the contact signature. The original energy formulation used in \cite{Zanfir_2018_CVPR} is given by
\begin{equation}
\label{eq:mp_loss}
  L =  \sum_{i \in \small\{ 1,2 \small\}}(L_{S}(P_i)+L_{psr}(P_i)) + L_{col}(P_1,P_2)
\end{equation}
where $L_{S}$ corresponds to the 2d semantic projection error with respect to the visual evidence extracted from the image (\ie semantic body part labeling and 2d pose) and $L_{psr}$ is a term for pose and shape regularization. $L_{col}(P_1,P_2)$ is a 3d collision penalty between $P_1$ and $P_2$ computed on a set of bounding sphere primitives. 
Gradients are passed from the loss function, through the 3d body model, all the way to the pose and shape parameters. Our body model has articulated body and hands \cite{xu2020ghum,MANO2017} and we additionally use estimated 2d hand joints positions.
This is necessary when modeling two interacting people, as hands are often involved in physical contact (see fig.~\ref{fig:clicks_frequencies}). We reflect these changes in the adapted terms $L^{\star}_{S}$ and $L^{\star}_{psr}$ and also introduce a new contact signature loss term, $L_{G}(P_1,P_2)$, that measures the geometric alignment of regions in correspondence. The adapted energy formulation becomes
\begin{align}
\label{eq:full_loss}
  L^{\star} = &\sum_{i \in \small\{ 1, 2 \small\}} (L^{\star}_{S}(P_i)+L^{\star}_{psr}(P_i)) + \\
  \nonumber
 &L_{col}(P_1,P_2) + L_{G}(P_1,P_2)
\end{align}
where 
\begin{align}
\label{eq:contact_loss}
    L_{G}(P_1,P_2)=L_{D}(P_1,P_2)+L_{N}(P_1,P_2)
\end{align}
The first term $L_{D}(P_1,P_2)$ seeks to minimize the sum of the distances between all the region pairs that are in contact, $(r_1, r_2) \in C(P_1,P_2)$
\begin{align}
    L_{D}(P_1,P_2) = \sum_{(r_1, r_2)\in C(P_1,P_2)} \Phi_D(r_1, r_2)
\end{align}
where the distance between two regions $\Phi_D(r_1, r_2)$ is
\begin{align}
\label{eq:PhiD}
    \Phi_D(r_1, r_2) = &\sum_{f_1\in \psi_D(r_1)}\min_{f_2 \in \psi_D(r_2)} \phi_D(f_1,f_2) + \\
    \nonumber
    &\sum_{f_2\in \psi_D(r_2)}\min_{f_1 \in \psi_D(r_1)} \phi_D(f_1,f_2)
\end{align}

\begin{table*}[!htbp]
\begin{center}
\scalebox{0.77}{
\begin{tabular}{|c|cc|cc|cc|cc|cc|cc|cc|cc|cc|}
\hline
\textbf{ } &
 \multicolumn{2}{|c|}{\textbf{Grab}} & \multicolumn{2}{|c|}{\textbf{Hit}} & \multicolumn{2}{|c|}{\textbf{Handshake}} & \multicolumn{2}{|c|}{\textbf{Holding hands}} & \multicolumn{2}{|c|}{\textbf{Hug}}   & \multicolumn{2}{|c|}{\textbf{Kick}} & \multicolumn{2}{|c|}{\textbf{Posing}} & \multicolumn{2}{|c|}{\textbf{Push}} & \multicolumn{2}{|c|}{\textbf{OVERALL}}\\ 

 \textbf{Optim.} & 
\multicolumn{1}{|c}{Pose } & \multicolumn{1}{c|}{Trans.} & \multicolumn{1}{|c}{Pose } & \multicolumn{1}{c|}{Trans. } &  \multicolumn{1}{|c}{Pose } & \multicolumn{1}{c|}{Trans. } &  \multicolumn{1}{|c}{Pose } & \multicolumn{1}{c|}{Trans. } &  \multicolumn{1}{|c}{Pose } & \multicolumn{1}{c|}{Trans. } &  \multicolumn{1}{|c}{Pose } & \multicolumn{1}{c|}{Trans. }  &  \multicolumn{1}{|c}{Pose } & \multicolumn{1}{c|}{Trans. }  &  \multicolumn{1}{|c}{Pose } & \multicolumn{1}{c|}{Trans. }  &  \multicolumn{1}{|c}{Pose } & \multicolumn{1}{c|}{Trans. }\\

 \textbf{Loss}& 
\multicolumn{2}{|c|}{Contact Dist.}  & \multicolumn{2}{|c|}{Contact Dist.}  &  \multicolumn{2}{|c|}{Contact Dist.}  & \multicolumn{2}{|c|}{Contact Dist.}  &  \multicolumn{2}{|c|}{Contact Dist.}  & \multicolumn{2}{|c|}{Contact Dist.}  &  \multicolumn{2}{|c|}{Contact Dist.}  & \multicolumn{2}{|c|}{Contact Dist.}  & \multicolumn{2}{|c|}{Contact Dist.}  \\
\hline
 \hline
\textbf{ $L^{\star}$ } &  
\multicolumn{1}{|c}{$\mathbf{116.5}$} & $\mathbf{390.14}$ & \multicolumn{1}{|c}{$\mathbf{119.4}$} &  $\mathbf{367.1}$ & \multicolumn{1}{|c}{$\mathbf{96.8}$} & $\mathbf{387.7}$ & \multicolumn{1}{|c}{$100.9$} & $\mathbf{379.5}$ & \multicolumn{1}{|c}{$\mathbf{173.9}$} & $\mathbf{400.2}$ & \multicolumn{1}{|c}{$\mathbf{140.0}$} & $\mathbf{419.2}$ & \multicolumn{1}{|c}{$\mathbf{138.8}$} & $\mathbf{364.3}$ & \multicolumn{1}{|c}{$\mathbf{116.9}$} & $\mathbf{380.5}$ 
& \multicolumn{1}{|c}{$\mathbf{125.4}$} & $\mathbf{368.0}$ \\

 &  \multicolumn{2}{|c|}{$\mathbf{19.1}$ ($\mathbf{3.5}$)} &
\multicolumn{2}{|c|}{$\mathbf{8.1}$ ($\mathbf{4.4}$)} &
\multicolumn{2}{|c|}{$\mathbf{12.1}$ ($\mathbf{2.8}$)} &
\multicolumn{2}{|c|}{$\mathbf{19.8}$ ($\mathbf{3.2}$)} &
\multicolumn{2}{|c|}{$\mathbf{62.0}$ ($\mathbf{44.5}$)} &
\multicolumn{2}{|c|}{$\mathbf{32.4}$ ($\mathbf{6.7}$)} &
\multicolumn{2}{|c|}{$\mathbf{40.8}$ ($\mathbf{10.9}$)} &
\multicolumn{2}{|c|}{$\mathbf{14.4}$ ($\mathbf{4.3}$)} &
\multicolumn{2}{|c|}{$\mathbf{26.0}$ ($\mathbf{10.0}$)} \\
 \hline
 \textbf{$L^{\star}$ w/o $L_G$} & 
 \multicolumn{1}{|c}{$121.1$} & $415.9$ &  
 \multicolumn{1}{|c}{$127.7$} & $395.7$ & 
 \multicolumn{1}{|c}{$98.5$} & $406.3$ & 
 \multicolumn{1}{|c}{$\mathbf{100.3}$} & $388.8$ & 
 \multicolumn{1}{|c}{$180.4$} & $424.4$ & 
 \multicolumn{1}{|c}{$154.8$} & $460.1$ & 
 \multicolumn{1}{|c}{$139.5$} & $376.9$ & 
 \multicolumn{1}{|c}{$123.6$}  & $399.4$ & 
 \multicolumn{1}{|c}{$130.7$}  & $408.4$\\
 
 \cite{Zanfir_2018_CVPR} &
 \multicolumn{2}{|c|}{$459.0$ ($366.3$)} &
 \multicolumn{2}{|c|}{$425.8$ ($363.4$)} &
 \multicolumn{2}{|c|}{$377.1$ ($305.2$)} &
 \multicolumn{2}{|c|}{$373.4$ ($273.9$)} &
 \multicolumn{2}{|c|}{$368.4$ ($327.5$)} &
 \multicolumn{2}{|c|}{$549.9$ ($464.2$)} &
 \multicolumn{2}{|c|}{$388.3$ ($327.0$)} &
 \multicolumn{2}{|c|}{$425.1$ ($369.4$)} &
 \multicolumn{2}{|c|}{$420.8$ ($349.6$)} \\

 \hline
\end{tabular}
}
\end{center}
\caption{3d human \textbf{pose} and \textbf{translation} estimation errors, as well as mean (median) 3D \textbf{contact distance}, expressed in mm, for the \chidb{} dataset. Our full optimization function, with the geometric alignment term on contact signatures, decreases the pose and translation estimation errors and the 3D distance between the surfaces annotated to be in contact. Higher contact distances are noticeable for complex interactions with complex contact signatures, such as hugging. As the parameters of our method are validated to minimize primarily pose reconstruction error, we do not necessarily achieve $0$ contact distance. This can be more tightly enforced by increasing the importance of the geometric alignment term, $L_G$, in the energy formulation, at the expense of a slightly increased reconstruction error.}
\label{table:results_reconstruction}
\end{table*}
For a facet in one region, this function takes its respective first nearest neighbor facet in the second region and computes the Euclidean distance $\phi_D(f_1,f_2)$ between the centers of the two facets, $f_1$ and $f_2$. Our approach is similar to iterative closest point, but performed at facet level.  For each region, we consider a subset of facets obtained by applying a selection operator $\psi_D$. 
This offers flexibility to operate not only on the entire set of facets (computationally intensive), but also on a fixed number of uniformly sampled facets or on a given subset of facets, \eg in the case of ground-truth facet level correspondences.

The second term, $L_{N}(P_1,P_2)$, enforces the orientation alignment for all region surfaces in contact, $(r_1, r_2) \in C(P_1,P_2)$
\begin{equation}
    L_{N}(P_1,P_2) = \sum_{(r_1, r_2)\in C(P_1,P_2)} \Phi_N(r_1, r_2)
\end{equation}
where $\Phi_N(r_1, r_2)$ measures the orientation alignment of a correspondence as the sum of all the orientation alignments between \textit{selected} pairs of facets from $r_1$ and $r_2$
\begin{equation}
\Phi_N(r_1, r_2) = {\sum_{(f_1, f_2) \in \psi_N(r_1, r_2)}\phi_N(f_1,f_2)}
\end{equation}
Here, the selection operator $\psi_N$ can re-utilize the facet level matches found in \eqref{eq:PhiD} or other defined facet level correspondences. To construct $\phi_N$, we start by defining the normal of facet $f=(v_1,v_2,v_3)$ as the cross product of its sides
\begin{equation}
    N(f) = (v_2 - v_1)\times(v_3-v_1) 
\end{equation}
This normal vector always points %
\textit{outside} %
the body by convention. The normal vector $N(f)$ has unit norm, $ \overline{N(f)} = N(f)/\left \| N(f) \right \| $. We align two facets such that their normal vectors are opposite (\ie parallel and of different sign)
\begin{equation}
    \phi_N(f_1,f_2) = 1+\overline{N(f_1)}\bullet \overline{N(f_2)}
\end{equation}

\subsection{3D Reconstruction in a Controlled Setup - \chidb{} \label{sec:chi3d_gt}}
Obtaining the ground-truth pose and shape of two interacting people is non-trivial even in controlled environments. Current technologies, if able to motion-capture multiple people at the same time, function properly when the subjects are not in close proximity and not occluding one another. The MoCap system available to us when recording \chidb{} is designed to track the 3d positions of the markers on only one person at a time. Also, the placement configuration of the markers on the body surface does not allow tracking most of the hand articulation. This motivates us to leverage a multitude of additional information: multi-view keypoint detections, scans of the subjects, annotated contact intervals and signatures, body pose priors, collision costs. 

Each of the subjects is scanned and accurate shape parameters $\bbeta$ are obtained by fitting the GHUM model to the scans, as in \cite{xu2020ghum}. 
For each sequence, we fit the GHUM \cite{xu2020ghum} parametric model to each of the subjects, starting with the participant wearing the markers.
The shape parameters $\bbeta$ are kept fixed during the pose $\btheta$, global rotation $\mathbf{R}$ and global translation $\mathbf{T}$ optimization, which is performed one frame at a time. Initialization for the optimization is performed in the first frame of each sequence by running THUNDR \cite{zanfir2021thundr} in each view and averaging the results. Optimization on all following frames is initialized from the result of the optimization in the respective previous frame.

We use a standard \textbf{3d marker alignment loss}, $\mathcal{L}_{3D}$, that measures the error between the 3d markers from the GHUM reconstructions $\left\{ \mathbf{m}^\text{3d}_{i} \right\}_{i=1, \ldots, K}$ and the available 3d ground truth markers $\left\{ \mathbf{M}^\text{3d}_{i} \right\}_{i=1, \ldots, K}$:
\begin{align}
    \mathcal{L}_{3d} = \frac{1}{K}\sum_{i=1}^{K}\| \mathbf{m}^\text{3d}_{i} -  \mathbf{M}^\text{3d}_{i}\|_{2}^{2}.
\end{align}
This loss is available only for the subject motion tracked by the VICON system.

We also use the standard \textbf{2d keypoint alignment loss}, $\mathcal{L}_{2d}$ in each of the $4$ RGB views of the scene, between the 3d joints regressed from the GHUM reconstructions and then projected to the image plane, $\left\{ \mathbf{j}^\text{2d}_{i} \right\}_{i=1, \ldots, K}$, and the 2d keypoints estimated from the image, $\left\{ \mathbf{J}^\text{2d}_{i} \right\}_{i=1, \ldots, K}$.
\begin{align}
    \mathcal{L}_{2d} = \frac{1}{K}\sum_{i=1}^{K}\| \mathbf{j}^\text{2d}_{i} -  \mathbf{J}^\text{2d}_{i}\|_{2}^{2}.
\end{align}
For the subject wearing markers, 2d keypoints are estimated only on a tight crop around the subject. We obtain a clean crop by projecting the 3d ground-truth %
markers to the image with the known camera parameters and cropping around their bounding box. For the subject not wearing markers, we run a keypoint predictor on the whole image. For each keypoint type, out of the several detections, we first drop the ones in close proximity to the keypoint of the subject wearing markers. From the remaining ones, we select the detection with the highest confidence. For the subject wearing markers, this $\mathcal{L}_{2d}$ loss is activated only for the hand keypoints, where not enough markers are present. For the other subjects, this loss is computed over all keypoint types.

To prevent self-intersections, we use the \textbf{self-collision loss} defined in \cite{fieraruNeurIPS2022}. For a GHUM mesh $\MM=(\VV, \FF)$ with vertices $\VV$ and faces $\FF$, we compute its PHUM \cite{fieraruNeurIPS2022} approximation $\MMH$. We apply the generalized winding number test \cite{jacobson2013robust} $L = \phi_{\MMH}(\VVH)$ on its set of vertices $\VVH$, with respect to its own mesh topology $\MMH$ and gather the binary inside/outside labeling $L \in \{0, 1\}^{N_{\hat{v}}\times 1}$. The vertices marked as inside, $\VVH_{L_{+}}$, with $L_{+} = \{l \in L: l = 1 \}$, are pushed out of the mesh in the direction of their nearest neighbor vertices, $\text{NN}(\hat{\VV}_{L_{+}}, \MMH)$:
\begin{equation}
    \mathcal{L}_{sc}=\sum_{l\in L_{+}}\|\VVH_l - \text{NN}(\hat{\VV}_{l}, \MMH)\|_{2}.
\end{equation}

We also adopt the \textbf{interpenetration loss} from  \cite{fieraruNeurIPS2022}. Given a pair of GHUM meshes $\MM_1$ and $\MM_2$ we compute their PHUM approximations $\MMH_1 = (\VVH_1, \FFH)$ and $\MMH_2 = (\VVH_2, \FFH)$ %
and the corresponding generalized winding number \cite{jacobson2013robust} tests for both $L_1 = \phi_{\MMH_2}(\VVH_1)$ and $L_2 = \phi_{\MMH_1}(\VVH_2)$. The loss is then given by:
\begin{align}
    \mathcal{L}_{ic}= &\sum_{l\in L_1{+}}\|\VVH_{1,l} - \text{NN}(\hat{\VV}_{1,l}, \MMH_2)\|_{2} +  \\
    \nonumber
    &\sum_{l\in L_2{+}}\|\VVH_{2,l} - \text{NN}(\hat{\VV}_{2,l}, \MMH_1)\|_{2}
\end{align}
In order to compute the nearest neighbor for inside vertices of a mesh, all vertices in the colliding mesh are queried. Since fitting is sequential, we only enable $\mathcal{L}_{ic}$ when fitting the second person. 

We employ a \textbf{ground plane contact loss}. For a given GHUM mesh $\mathbf{M} =(\mathbf{V}, \mathbf{F})$ we define the top $k$ ground plane penetrating distances $\mathbf{D}_{k} = \mathbf{top}_{k} \left( \max\left( 0, -d\left( \mathbf{V}, \mathcal{T} \right) \right) \right)$ where $d\left( \mathbf{V}, \mathcal{T} \right)$ are the signed distances from the GHUM mesh vertices $\mathbf{V}$ and the ground plane $\mathcal{T}$. Then $\mathcal{L}_{gp} = \displaystyle \sum_{x \in \mathbf{D}_{k}} \| x \|_{2}$.

For the \textbf{pose prior}, we use regularization loss on the normalizing flow ($\text{NF}$) latent code introduced in \cite{zanfir2020weakly}:  $\mathcal{L}_{\theta} = \|\text{NF}(\btheta)\|_2^2$. 

When fitting the second person, we also employ a \textbf{smoothing loss} $\mathcal{L}_{s} = \|\btheta_{t} - \btheta_{t-1}\|_2^2 + \|\mathbf{R}_{t} - \mathbf{R}_{t-1}\|_2^2 + \|\mathbf{T}_{t} - \mathbf{T}_{t-1}\|_2^2$ where $\btheta_{t}$ are the pose variables for the $t^\text{th}$ frame.

When optimizing the second person, to bring it in contact with the one wearing markers, we add the \textbf{contact signature penalty term} $L_{G}$ defined in eq.~\ref{eq:contact_loss} using ground truth facet correspondences. Since the contact time interval is fully annotated, %
we use this penalty term in all contact frames.

\begin{figure*}
[!htbp]
\def\wf{.27}
\setlength{\tabcolsep}{6pt}
\begin{center}
\begin{tabular}{cccc}
\rotatebox{90}{\hspace{16pt}\textit{Two people are hugging.}} &
\includegraphics[width=\wf\linewidth]{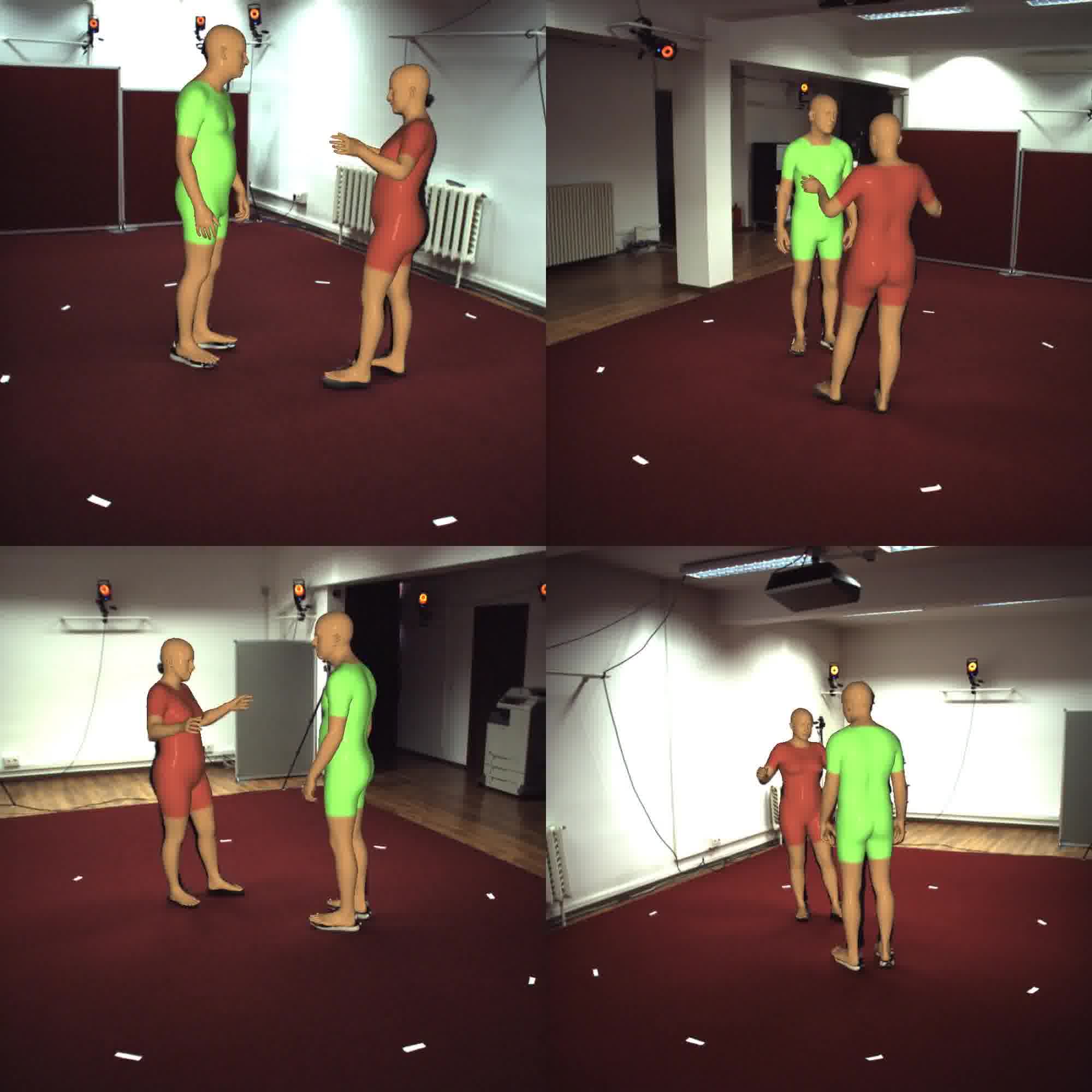} & 
\includegraphics[width=\wf\linewidth]{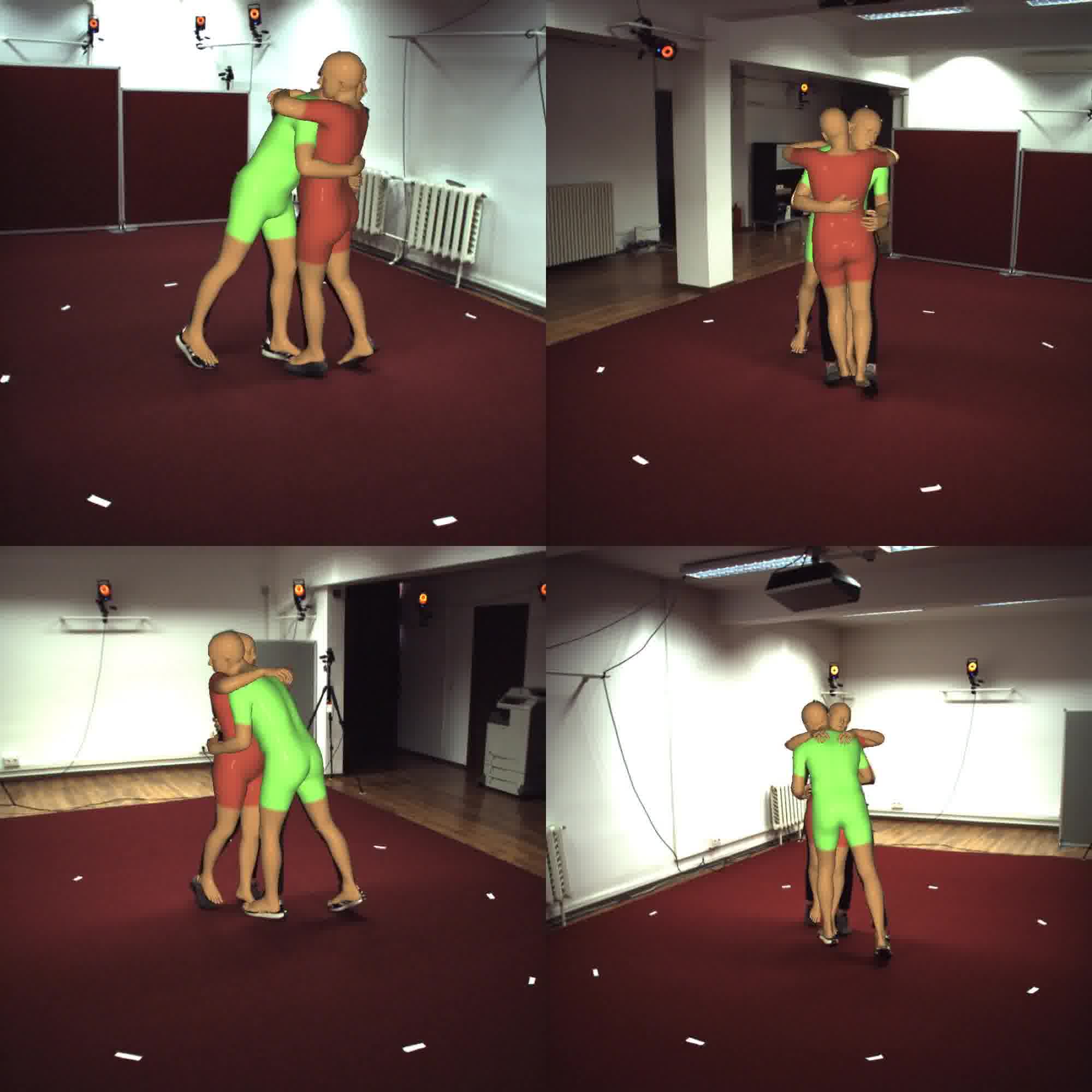}  &
\includegraphics[width=\wf\linewidth]{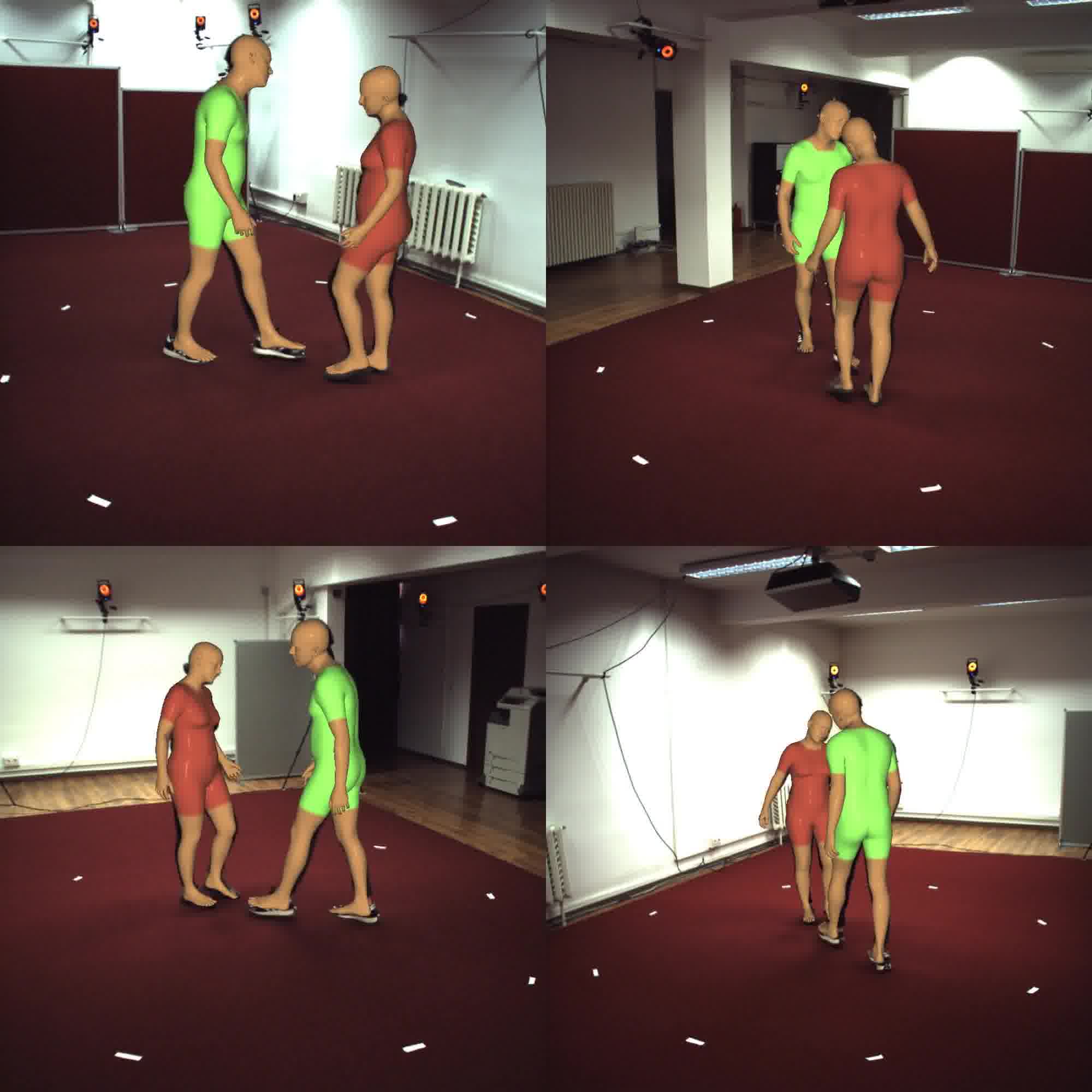} \\
\rotatebox{90}{\hspace{13pt}\shortstack{\textit{A man holds his right}\\ \textit{arm around somebody's} \\\textit{shoulder and raises his left}\\ \textit{hand for a picture.}}} &
\includegraphics[width=\wf\linewidth]{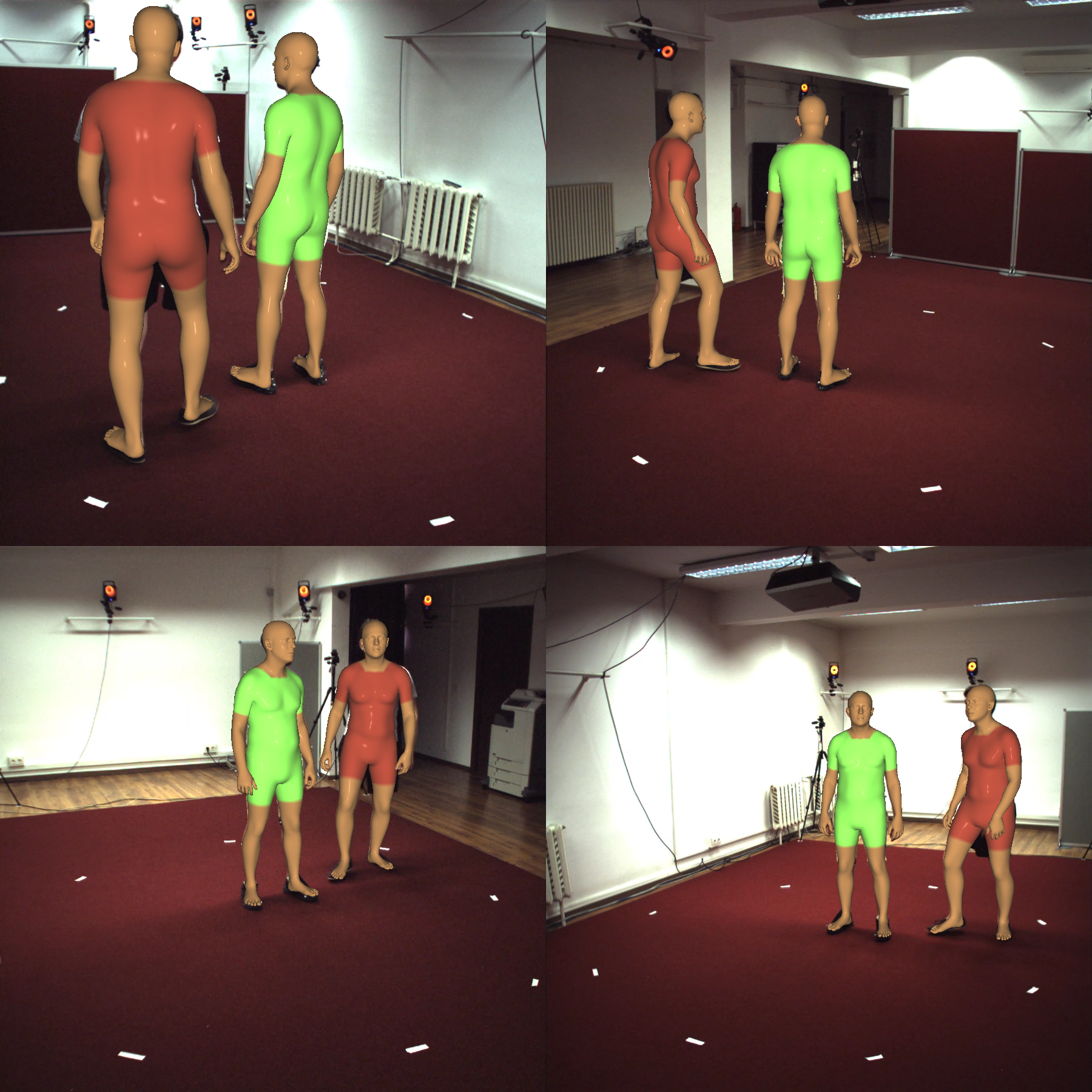} & 
\includegraphics[width=\wf\linewidth]{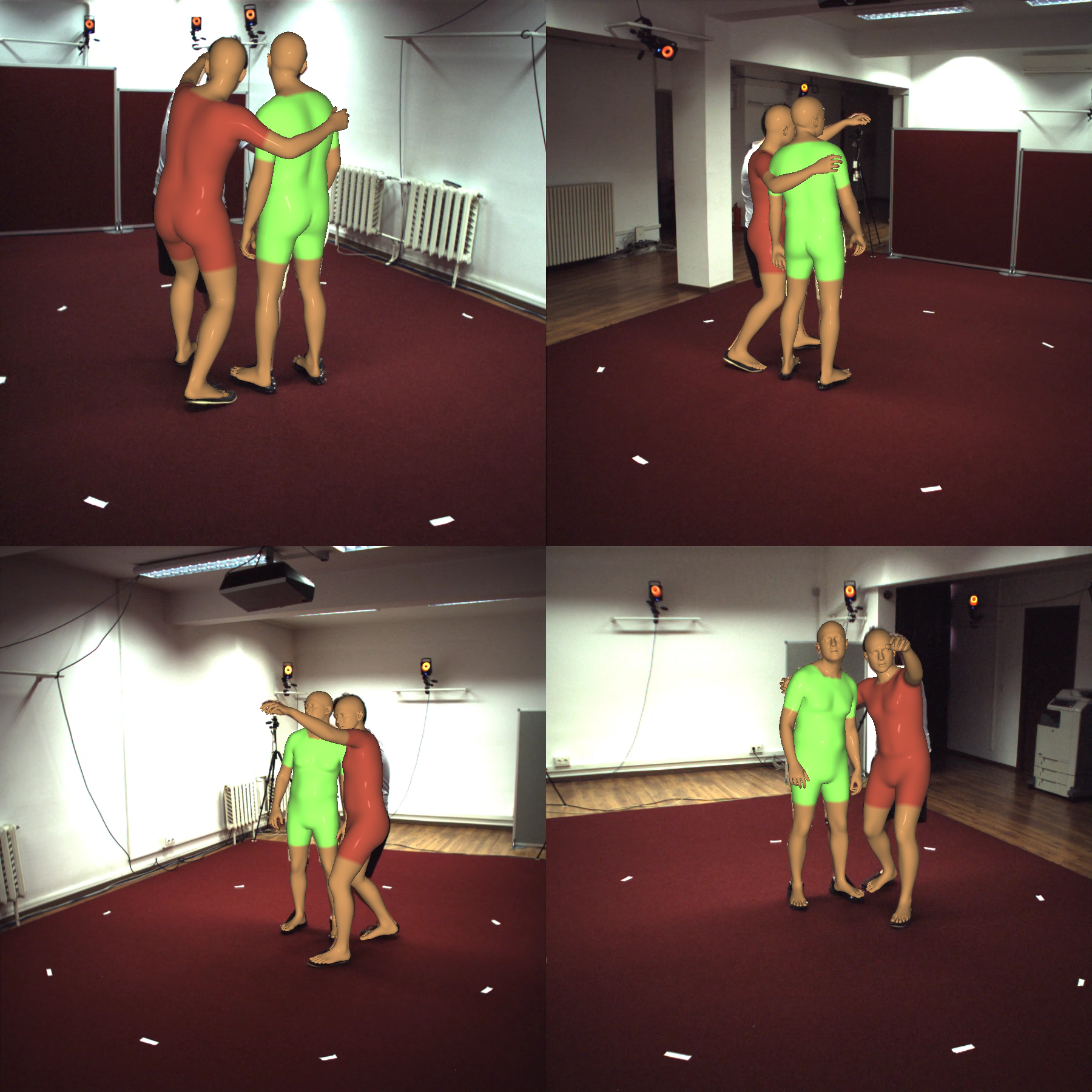} &
\includegraphics[width=\wf\linewidth]{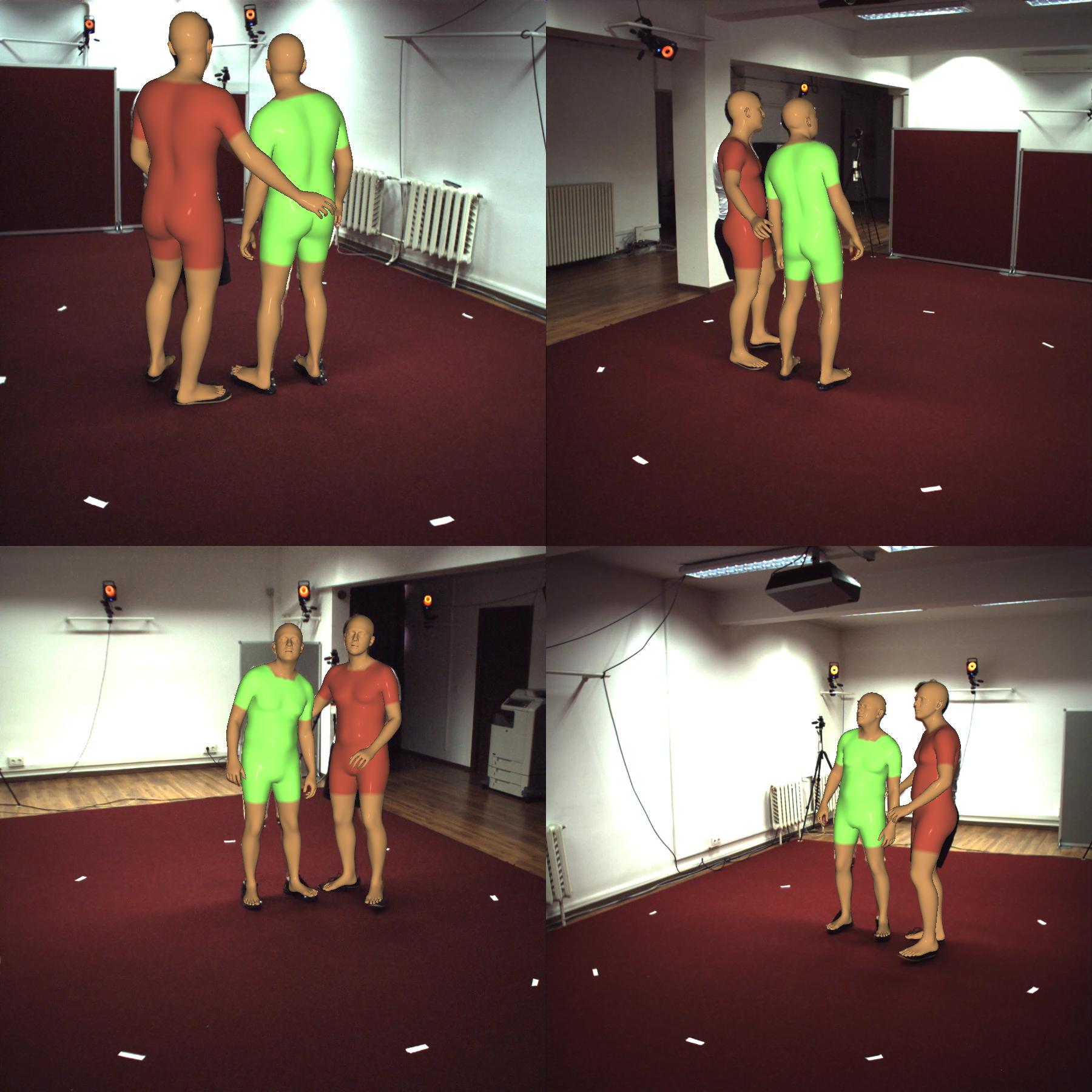}
                     
\end{tabular}
\end{center}
\caption{\textit{First Column:} Annotated text describing the motion. \textit{Column 2-4:} Fitting of the GHUM model to two interacting subjects in the \chidb{} dataset. All 4 stacked views are displayed for 3 frames (left to right in temporal order): a hugging sequence (top), a posing sequence (bottom). Subject in green is the subject wearing the markers. Subject in brick-red does not wear any markers on clothing.}
\label{fig:chi3d_fitting}
\end{figure*}

The final loss is a weighted sum of all individual losses. All 3d reconstructions in the CHI3D dataset were visually inspected from all viewpoints during several stages of the design of the method, motivating all losses introduced. Manual adjustment of the 3d markers on the GHUM template was performed for each subject, as the marker placement during capture did vary slightly from subject to subject. Visualization results for GHUM fitting the interacting subjects in \chidb{} are shown in Fig.~\ref{fig:chi3d_fitting}. 

In addition, we also convert the GHUM sequences to the SMPLX format and publicly release all motions is both formats for training and evaluation purposes.

\section{Experiments }
In the following section, we describe our experiments and results on contact classification and segmentation and signature prediction (sec.~\ref{sec:res_contacts}) and on monocular 3D reconstruction (sec.~\ref{sec:monocular_rec_res}). We also present the evaluation protocol and show benchmark results for recent multi-people reconstruction methods in sec.~\ref{sec:benchmark}.
\subsection{Contact-Based Tasks\label{sec:res_contacts}}
We report quantitative results on our collected FlickrCI3D dataset. We split both the contact classification database ($90,167$ images) and the contact segmentation and correspondences database ($14,081$ images) into train, validation and test subsets each, using the following proportions $85\%$, $7.5\%$ and $7.5\%$ respectively. In all our experiments, we validate the meta-parameters on the validation set and report the results on the test set.
\begin{figure*}
[!htbp]
\def\hf{67pt}
\begin{center}

         \includegraphics[height=\hf]{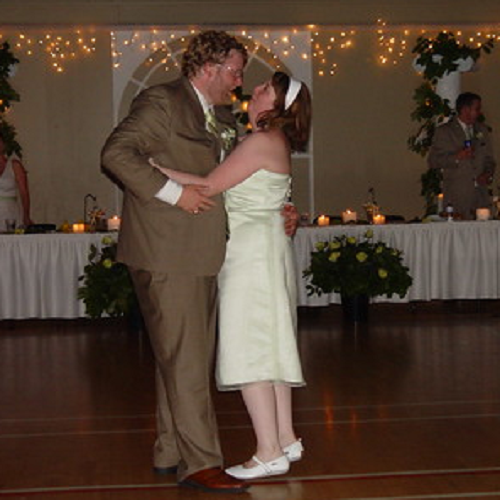}
         \includegraphics[height=\hf]{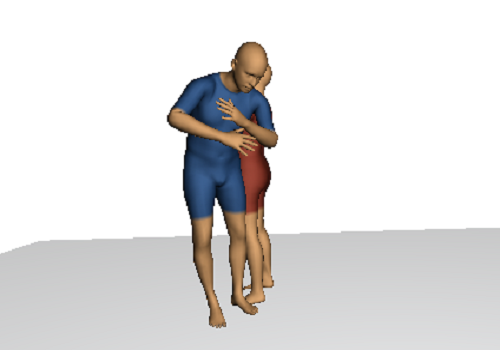}
         \includegraphics[height=\hf]{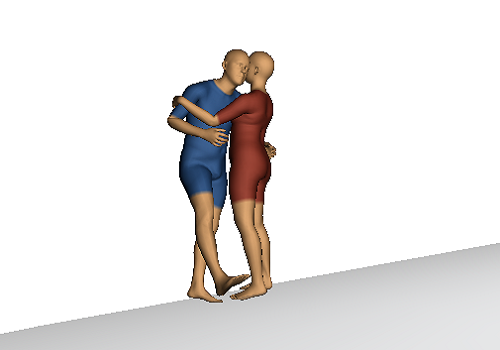}
         \includegraphics[height=\hf]{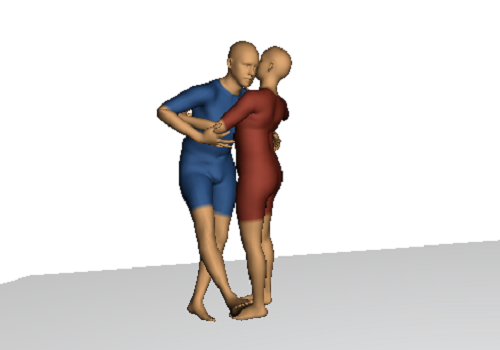}
         \includegraphics[height=\hf]{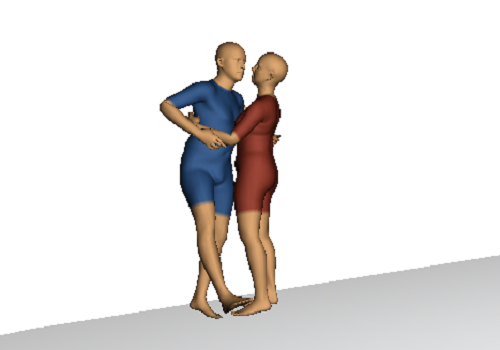}
         \\
         \includegraphics[height=\hf]{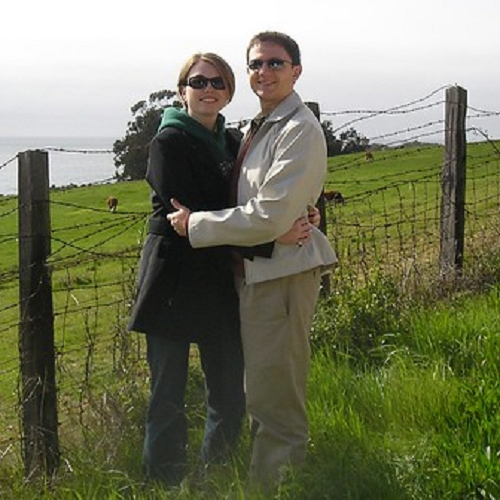}
         \includegraphics[height=\hf]{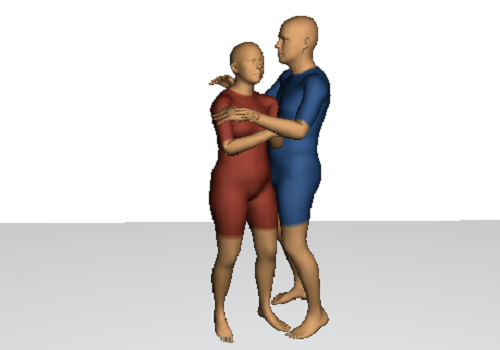}
         \includegraphics[height=\hf]{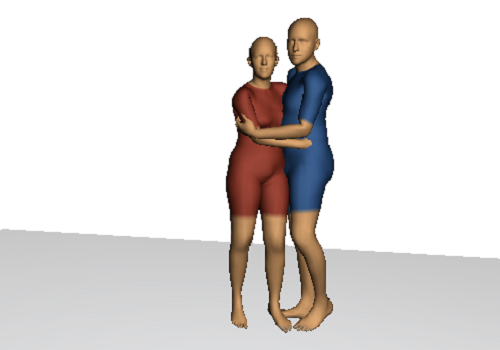}
         \includegraphics[height=\hf]{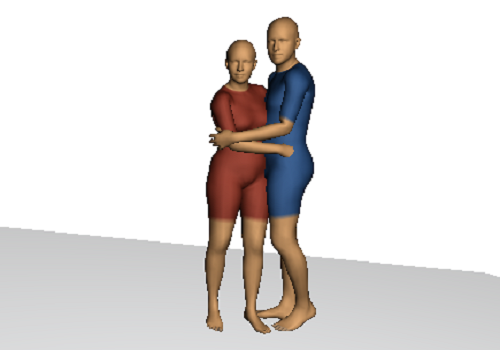}
         \includegraphics[height=\hf]{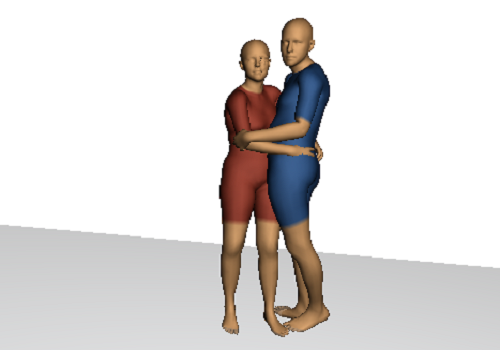}
         \\
         \includegraphics[height=\hf]{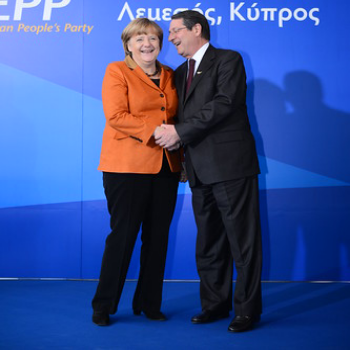}
         \includegraphics[height=\hf]{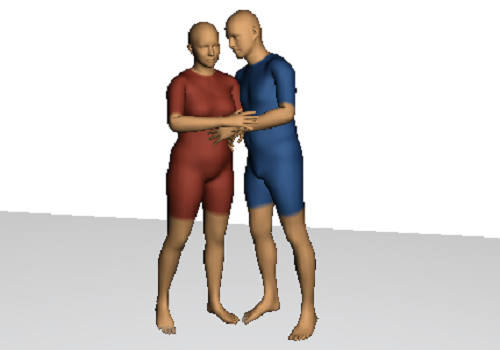}
         \includegraphics[height=\hf]{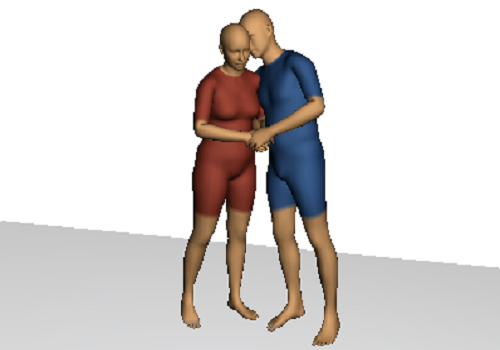}
         \includegraphics[height=\hf]{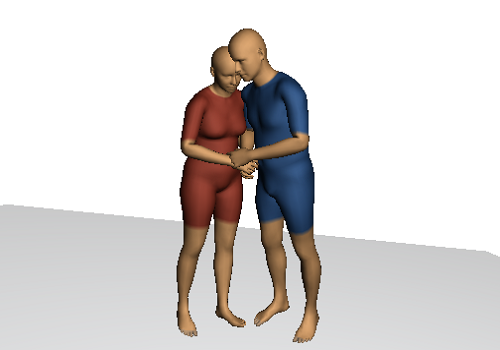}
         \includegraphics[height=\hf]{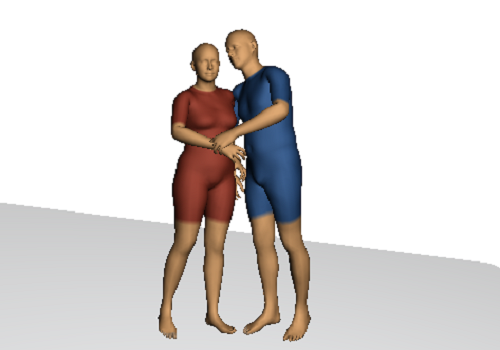}
         \\
         \includegraphics[height=\hf]{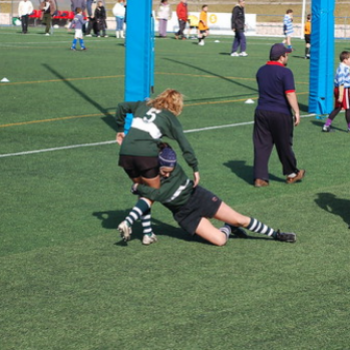}
         \includegraphics[height=\hf]{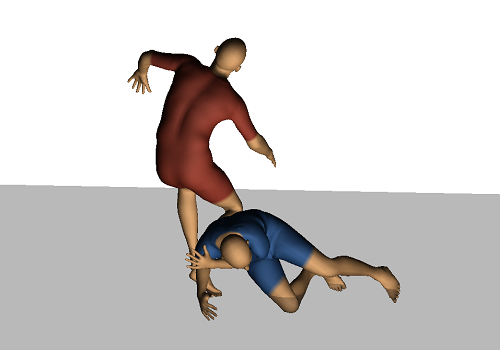}
         \includegraphics[height=\hf]{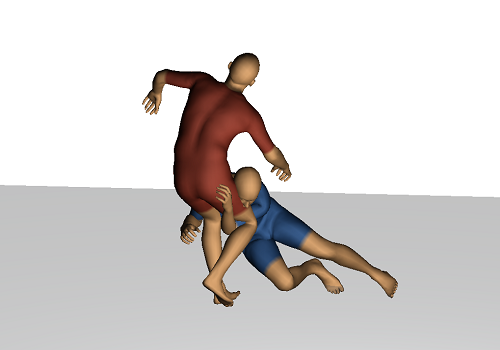}
         \includegraphics[height=\hf]{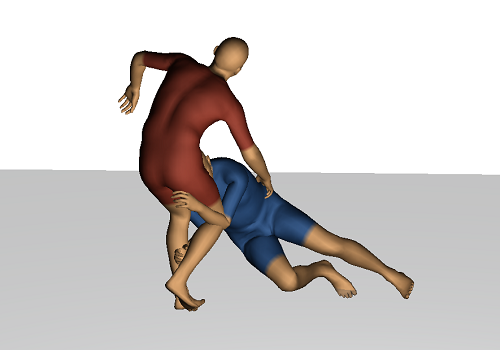}
         \includegraphics[height=\hf]{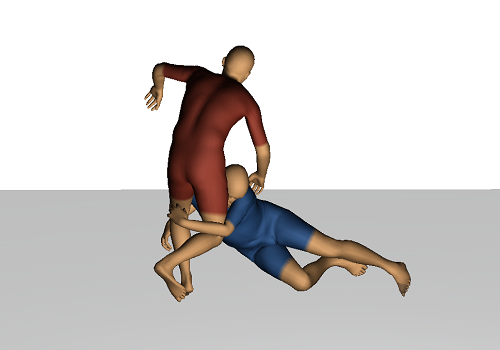}
         \\
         \includegraphics[height=\hf]{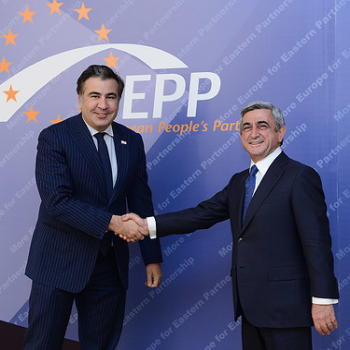}
         \includegraphics[height=\hf]{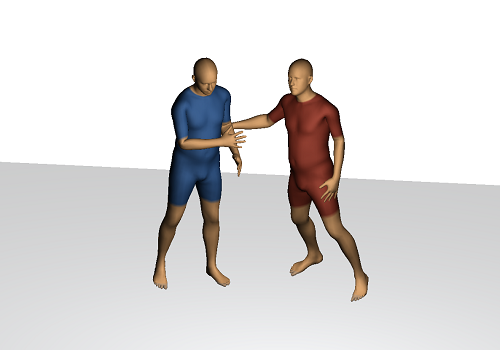}
         \includegraphics[height=\hf]{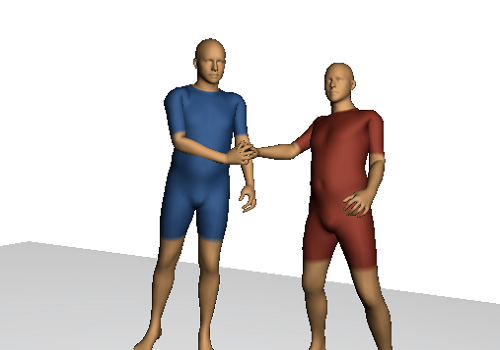}
         \includegraphics[height=\hf]{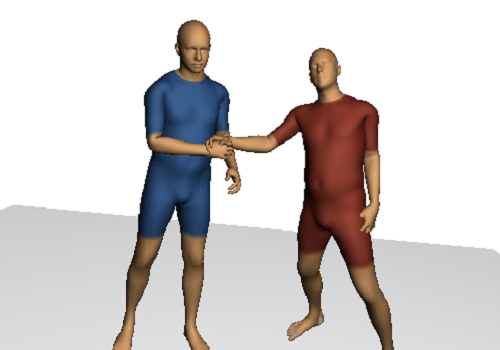}
         \includegraphics[height=\hf]{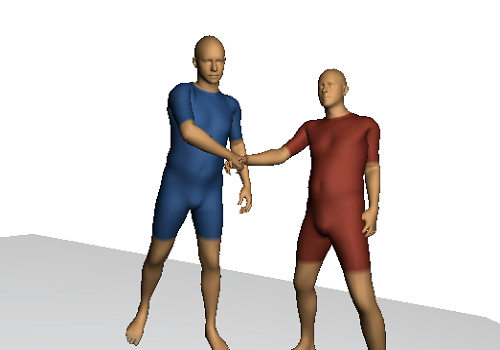}
\end{center}
\caption{3D human pose and shape reconstructions using contact constraints of different granularity. The first column shows the RGB images followed by their reconstructions without contact information (column 2), using contacts based on 37 and 75 regions, respectively (columns 3 \& 4), and using facet-based correspondences (column 5). While using facet-based constraints provides the most accurate estimates, reasonable results can be obtained even for coarser (region) assignments. }
\label{fig:flickr-reconstructions}
\end{figure*}
We evaluate the performance of the contact detection task and obtain an average accuracy of $0.846$, with  $0.844$ for the "contact" class and $0.848$ for the "no contact" class.

For the contact segmentation and signature prediction method, we train our network with $N_{reg}=75$, though we can also obtain the coarser versions post-hoc. In training, since our ground truth does not necessarily contain the full set of region correspondences, we do not penalize the non-annotated (but possible) correspondences between the segments on one person and those on the other. At inference time, we exploit
the contact segmentation and use its predictions to mask spurious correspondences.

We evaluate our predictions using the intersection over union ($\text{IoU}_{N_{reg}}$) metric, computed for different region granularities. Table~\ref{table:contact3d_quant} reports the performance of our full model, for which predictions get closer to the human performance as the region granularity becomes coarser. We also train a version of our method without concatenating the semantic 2d features to the input. In almost all cases, these input features affect performance positively. Similarly, jointly learning the two tasks and using the contact segmentation mask to eliminate non-valid correspondences improves the contact signature estimation performance. \\

\subsection{Monocular 3D Reconstruction Results\label{sec:monocular_rec_res}} In fig.~\ref{fig:flickr-reconstructions} we show reconstruction examples on images from FlickrCI3D using our proposed model formulation (see ~\eqref{eq:full_loss}) and contact annotations at different levels of granularity. The first column represents an RGB image from the FlickrCI3D dataset. %
Without contact information, the method of \cite{Zanfir_2018_CVPR} may provide plausible poses, sometimes with good image alignment, but the subtleties of the interaction are lost. In order to avoid mutual volume intersections, people may even be placed far from each other. By incorporating contact correspondences we can successfully recover the essential details of the interactions, such as hands touching during a handshake/dancing or tackling someone in a rugby match.

We also perform quantitative experiments on the \chidb{} dataset (on the frames where people are annotated to be in contact) and ablate the impact of our geometric alignment term, $L_G$. We select a small set of images ($160$) to validate the weight of each term in the energy function and report results on the remaining test set. We evaluate both the inferred pose, using mean per joint position error (MPJPE), and the estimated translation. 
In addition, on the estimated reconstructions, we compute the mean and median distance between the regions in the contact signature. This contact distance is defined as the minimum Euclidean distance between each pair of facets from two regions annotated to be in correspondence.
Results are given in table \ref{table:results_reconstruction} where annotated contact information improves the accuracy of the reconstruction. For pose, we only evaluate on a standard 3d body joints configuration compatible with the MoCap format that does not include body extremities (\eg hands and feet). Our complete optimization framework not only produces more accurate reconstructions of pose and translation, but also closely approaches the contact signature.

\subsection{Evaluation Protocol and Benchmark\label{sec:benchmark}}
To help improve the state of the art in 3D human reconstruction of close contact interactions, we propose a public challenge on the test set of the CHI3D dataset. We standardize the evaluation protocol and metrics to facilitate the comparison of different research methods.

Our evaluation server accepts multiple formats of the reconstructions: GHUM parameters, SMPLX parameters and 3D Joints (in the Human3.6M format with 17 joints). Evaluation is performed against both persons, where person matching between predictions and ground truth is performed per-frame, using 2d bounding boxes. Each 3d reconstruction method might use a different camera model, so the submitted reconstructions might project wrongly with the official camera model and parameters. We therefore ask submissions to also include a 2d bounding box for each reconstruction, which is used to perform the person matching using the Intersection over Union metric.
We require predictions only on a subset of frames, equally distanced. The released test set consists of only one random camera viewpoint per interaction scenario, so using multi-view triangulation in reconstruction is not possible. Each prediction has to be provided in the coordinate system of the provided camera. 

All evaluation metrics are averaged and reported over (1) all frames that were sampled, as well as only over the (2) contact frames. We test our monocular optimization-based method, as well as $3$ more recent learned methods. Ours and REMIPS \cite{fieraruNeurIPS2022} output GHUM predictions, which we also convert to SMPL-X, while CRMH \cite{jiang2020coherent} and Cliff \cite{li2022cliff} output SMPL meshes, which we convert to both SMPL-X and GHUM. Table~\ref{table:benchmark} shows that, although the recent learned methods have improved the state-of-the art over time, the reconstruction of interactions is still a challenging problem, with joints/vertex errors over $50$ millimeters.

\begin{table*}[!htbp]
\begin{center}
\begin{tabular}{|c|ccc|cc|cc|cc|}\hline
& \multicolumn{7}{c}{All Frames / Contact Frames} \vline \\
\hline
& \multicolumn{3}{c}{3D Joints} \vline &
\multicolumn{2}{c}{GHUM} \vline &
\multicolumn{2}{c}{SMPLX} \vline \\

\textbf{Method} 
& \textbf{Translation Error} & \textbf{MPJPE} & \textbf{MPJPE-PA} & \textbf{MPVPE}  & \textbf{MPVPE-PA} & \textbf{MPVPE} & \textbf{MPVPE-PA} \\
\hline
\hline
Ours (sec.~\ref{sec:monoc_rec}) & $764.17$ / $755.00$  & $89.68$ / $106.15$ & $59.95$ / $74.53$ & $101.34$ / $120.36$ & $70.25$ / $84.89$ & $93.04$ / $110.76$ & $58.71$ / $74.14$ \\
CRMH \cite{jiang2020coherent}& $3157.60$ / $3130.39$ & $92.21$ / $113.74$ & $54.83$ / $70.45$ & $106.47$ / $131.70$ & $65.11$ / $86.46$& $97.84$ / $119.74$ & $55.66$ / $73.58$ \\
REMIPS \cite{fieraruNeurIPS2022}& $676.17$ / $693.50$ & $84.05$ / $101.04$ & $55.85$ / $68.27$ & $96.02$ / $114.97$ & $65.01$ / $80.86$ & $87.08$ / $104.51$ & $54.87$ / $68.79$ \\
Cliff \cite{li2022cliff}& $836.53$ / $816.29$ & $72.37$ / $88.66$ & $47.50$ / $56.89$ & $73.60$ / $92.24$ & $49.44$ / $62.82$ & $67.01$ / $83.66$ & $42.51$ / $54.40$ \\
\hline
\end{tabular}
\end{center}
\caption{Public benchmark on 3D human reconstruction of close contact interactions on \chidb{}. Each metric (expressed in millimeters) is averaged either over all frames (left) or over the contact frames (right). Note that, although using the \chidb{} training set is not prohibited, the recent learned methods we benchmarked \cite{jiang2020coherent, fieraruNeurIPS2022, li2022cliff} were trained on other data sources.} 
\label{table:benchmark}
\end{table*}
\section{Conclusions}
We have argued that progress in human sensing and scene understanding would eventually require the detailed 3d reconstruction of human interactions where contact plays a major role, not only for veridical estimates, but in order to ultimately understand fine-grained actions, behavior and intent. We have proposed a graded modeling framework for Interaction Signature Prediction (ISP) based on contact detection and 3d correspondence estimation over model surface regions at different levels of detail, with subsequent 3d reconstruction under losses that integrate contact and surface normal alignment constraints. We have undertaken a major effort to collect 3d ground truth data of humans involved in interactions (CHI3D, 631 sequences containing 2,525 contact events, 728,664 ground truth reconstructions, 631 motion textual descriptions), as well as image annotations in the wild (FlickrCI3D, a dataset of $11,216$ images, with $14,081$ processed pairs of people, and $81,233$ facet-level surface correspondences within $138,213$ selected regions). We have evaluated all components in detail, showing their relevance towards accurate 3d reconstruction of human contact. Data in multiple formats (GHUM and SMPLX parameters, Human3.6m 3d joints) is made available for research purposes at https://ci3d.imar.ro, together with an evaluation server and a public benchmark.\\

\ifCLASSOPTIONcompsoc
  \section*{Acknowledgments}
\else
  \section*{Acknowledgment}
\fi

This work was supported in part by the ERC Consolidator grant SEED, CNCS-UEFISCDI (PN-III-P4-PCE-2021-1959) and SSF.

\ifCLASSOPTIONcaptionsoff
  \newpage
\fi

\bibliographystyle{IEEEtran}
\bibliography{human.bib}

\begin{IEEEbiography}[{\includegraphics[width=1in,height=1.25in,clip,keepaspectratio]{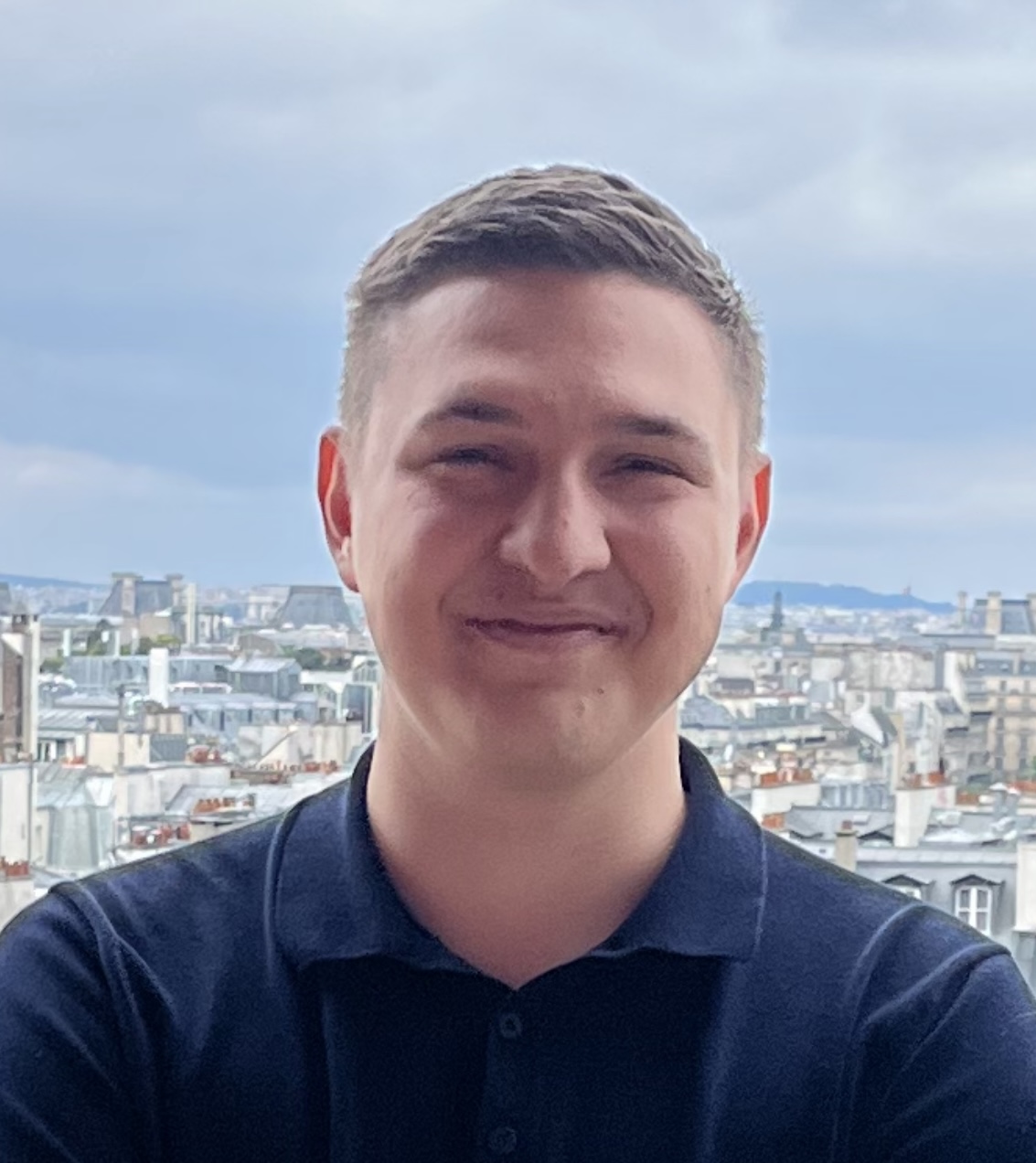}}]{Mihai Fieraru} is a PhD candidate at the School of Advanced Studies of the Romanian Academy. He holds a BSc. from Jacobs University Bremen and a MSc. from Saarland University, both in Computer Science. During his undergraduate studies, he spent one exchange semester at Carnegie Mellon University. He received an IMPRS-CS scholarship and fellowship from the Max Planck Institute for Informatics for the duration of his masters. His current interests are human understanding in images and videos, self-supervised learning and weakly-supervised learning.
\end{IEEEbiography}

\begin{IEEEbiography}[{\includegraphics[width=1in,height=1.25in,clip,keepaspectratio]{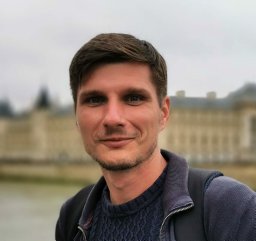}}]{Mihai Zanfir}
holds a BSc in Computer Science and a MSc in Artificial Intelligence from the University Politehnica of Bucharest and a PhD in Computer Science from the School of Advanced Studies of the Romanian Academy. During his PhD, his principal research interests were human action recognition from RGB and depth sensors, and 3D human sensing for the task of pose, shape and clothing reconstruction from natural images. He was also part of the DE-ENIGMA project, which developed artificial intelligence for a commercial robot in assisted therapy of children diagnosed with autism.
\end{IEEEbiography}

\begin{IEEEbiography}[{\includegraphics[width=1in,height=1.25in,clip,keepaspectratio]{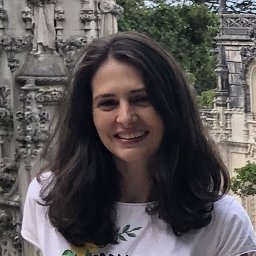}}]{Elisabeta Oneata}
holds a MSc degree in Artificial Intelligence from the University of Manchester and a PhD degree in Computer Science from the School of Advanced Studies of the Romanian Academy. She also did a research internship at the Bioinformatics Institute A*Star, Singapore. Her PhD studies focused on computer vision and machine learning, more particularly, the automated human understanding from monocular images: 2d/3d human pose and shape reconstruction, as well as human pose perception. She was also actively involved in the EU-Horizon project DE-ENIGMA, which developed the artificial intelligence for a robot used for on emotion-recognition and emotion-expression teaching programme to school-aged autistic children.
\end{IEEEbiography}

\begin{IEEEbiography}[{\includegraphics[width=1in,height=1.25in,clip,keepaspectratio]{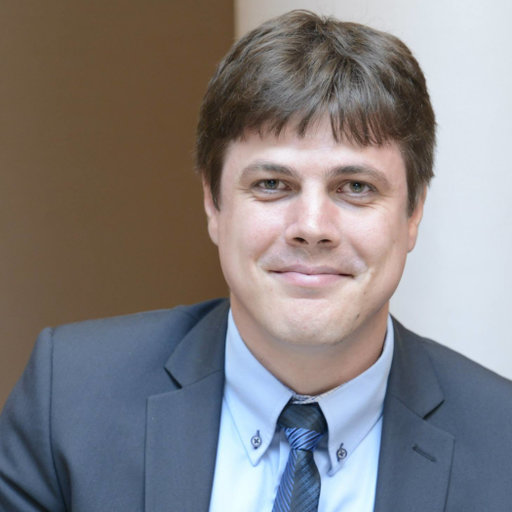}}]{Alin-Ionut Popa}
holds a MSc degree in Advanced Computing from the Faculty of Engineering, Imperial College London and a PhD degree in Computer Science from the School of Advanced Studies of the Romanian Academy. During his MSc, he obtained the "Open Horizons" Scholarship and did a research internship at the Bioinformatics Institute A*Star, Singapore. His PhD studies focused on computer vision and machine learning, more particularly, the problem of 3d human sensing from visual data.
\end{IEEEbiography}

\begin{IEEEbiography}[{\includegraphics[width=1in,height=1.25in,clip,keepaspectratio]{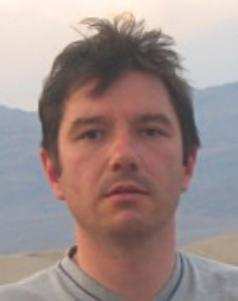}}]{Vlad Olaru}
holds a BS degree from the ”Politehnica” University of Bucharest, Romania, an MS degree from Rutgers, The State University of New Jersey, USA and a PhD from the Technical University of Karlsruhe, Germany. His research interests focus on distributed and parallel computing, operating systems, real-time embedded systems and high-performance computing for large-scale computer vision programs. His doctorate concentrated on developing kernel-level, single system image services for clusters of computers. He was a key person in several EU-funded as well as national projects targeting the development of real-time OS software to control the next generation of 3D intelligent sensors, real-time Java for multi-core architectures, robot-assisted therapy for autistic children.
\end{IEEEbiography}

\begin{IEEEbiography}[{\includegraphics[width=1in,height=1.25in,clip,keepaspectratio]{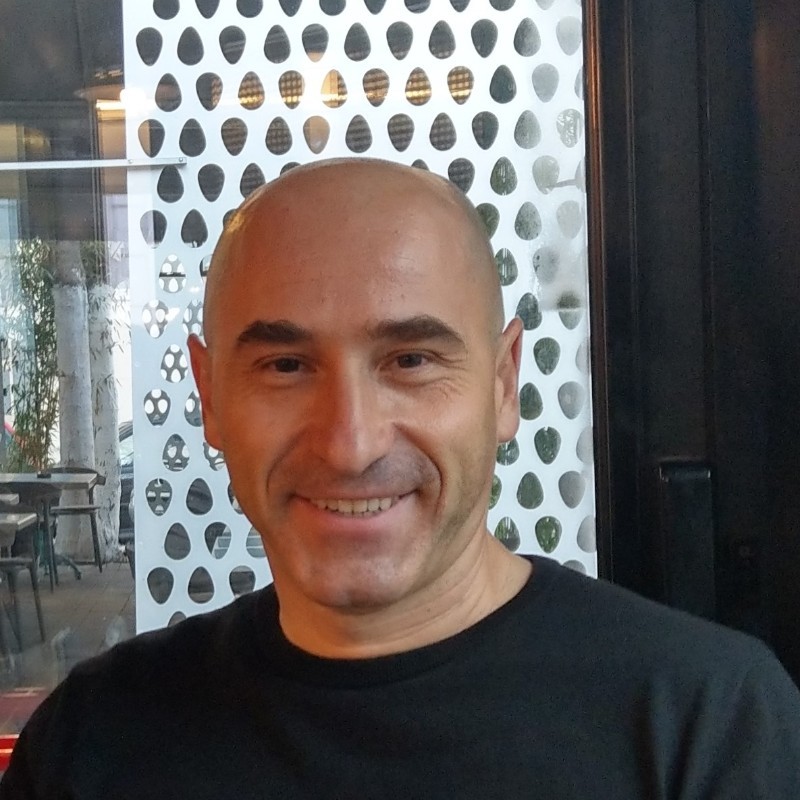}}]{Cristian Sminchisescu}
has obtained a doctorate in Computer Science and Applied Mathematics with an emphasis on imaging, vision and robotics at INRIA, France, under an Eiffel excellence doctoral fellowship, and has done postdoctoral research in the Artificial Intelligence Laboratory at the University of Toronto. He is a member in the program committees of the main conferences in computer vision and machine learning (CVPR, ICCV, ECCV, NIPS, AISTATS), area chair for ICCV07-13, and an Associate Editor of IEEE PAMI. He has given more than 100 invited talks and presentations and has offered tutorials on 3d tracking, recognition and optimization at ICCV and CVPR, the Chicago Machine Learning Summer School, the AERFAI Vision School in Barcelona and the Computer Vision Summer School (VSS) in Zurich. His research interests are in the area of computer vision (3D human pose estimation, semantic segmentation) and machine learning (optimization and sampling algorithms, structured prediction, and kernel methods).
\end{IEEEbiography}

\end{document}